\PassOptionsToPackage{table}{xcolor}
\documentclass[11pt]{article}
\usepackage[final]{acl}
\usepackage{times}
\usepackage{latexsym}
\usepackage[T1]{fontenc}
\usepackage[utf8]{inputenc}
\usepackage{microtype}
\usepackage{inconsolata}
\usepackage{graphicx}
\usepackage{amsmath}
\usepackage{amssymb}
\usepackage{amsthm}
\usepackage{booktabs}
\usepackage{multirow}
\usepackage{algorithm}
\usepackage{algpseudocode}
\usepackage{xcolor}
\usepackage{enumitem}
\usepackage[most]{tcolorbox}
\usepackage{varwidth}
\definecolor{bggray}{RGB}{217, 217, 217}
\usepackage[most]{tcolorbox}
\usepackage{xcolor}

\definecolor{PromptBG}{RGB}{250,250,248}
\definecolor{PromptBorder}{RGB}{120,125,132}
\definecolor{PromptHeader}{RGB}{50,58,69}
\definecolor{InnerBG}{RGB}{242,245,248}
\usepackage[most]{tcolorbox}
\tcbuselibrary{listings} 

\definecolor{InnerBG}{RGB}{242,245,248}
\definecolor{PromptBorder}{RGB}{120,125,132}
\definecolor{jsonKey}{RGB}{163, 21, 21}      
\definecolor{jsonValue}{RGB}{0, 0, 0}       
\definecolor{jsonNum}{RGB}{9, 134, 115}      
\definecolor{jsonComment}{RGB}{0, 128, 0}    

\lstdefinelanguage{myjson}{
  basicstyle=\small\ttfamily\color{PromptBorder!80!black},
  string=[s]{"}{"},
  stringstyle=\color{jsonKey},
  comment=[l]{//},
  commentstyle=\color{jsonComment},
  literate=
   *{0}{{{\color{jsonNum}0}}}{1}
    {1}{{{\color{jsonNum}1}}}{1}
    {2}{{{\color{jsonNum}2}}}{1}
    {3}{{{\color{jsonNum}3}}}{1}
    {4}{{{\color{jsonNum}4}}}{1}
    {5}{{{\color{jsonNum}5}}}{1}
    {6}{{{\color{jsonNum}6}}}{1}
    {7}{{{\color{jsonNum}7}}}{1}
    {8}{{{\color{jsonNum}8}}}{1}
    {9}{{{\color{jsonNum}9}}}{1},
  moredelim=*[s][\color{jsonValue}]{:\ "}{"},
  moredelim=*[s][\color{jsonValue}]{:\ \ "}{"},
  breaklines=true,
  showstringspaces=false
}

\newtcblisting{innerjson}{
  colback=InnerBG,
  colframe=PromptBorder!40,
  boxrule=0.5pt,
  arc=1.2mm,
  left=6pt,
  right=6pt,
  top=5pt,
  bottom=5pt,
  before skip=6pt,
  after skip=6pt,
  listing only,                           
  listing options={language=myjson}        
}

\newtcolorbox{promptcard}[1]{
  colback=PromptBG,
  colframe=PromptBorder,
  boxrule=0.8pt,
  arc=1.2mm,
  width=\linewidth,
  left=10pt,
  right=10pt,
  top=8pt,
  bottom=8pt,
  title=\textbf{#1},
  colbacktitle=PromptHeader,
  coltitle=white,
  fonttitle=\small,
  before skip=10pt, 
  after skip=10pt,
  sharp corners=south,
  rounded corners=northwest,
  rounded corners=northeast
}

\newtcolorbox{innerprompt}{
  colback=InnerBG,
  colframe=PromptBorder!40,
  boxrule=0.5pt,
  arc=1mm,
  left=6pt,
  right=6pt,
  top=5pt,
  bottom=5pt,
  before skip=6pt,
  after skip=6pt
}

\newtheorem{lemma*}{Lemma}

\newtheorem{proposition}{Proposition}[section]

\title{Skill-Conditioned Gated Self-Distillation for LLM Reasoning}
\author{Jiazhen Huang$^1$, Xiao Chen$^1$, Xiao Luo$^2$, Yong Dai$^3$, Senkang Hu$^4$, Yuzhi Zhao$^5$\\
$^1$Tsinghua University, $^2$University of Wisconsin-Madison, $^3$X-Humanoid, \\$^4$City University of Hong Kong, $^5$Huazhong University of Science and Technology\\
\texttt{huangjiazhen1125@gmail.com}}

\def\method{SGSD}

\begin{document}
\maketitle

\begin{abstract}

On-policy self-distillation (SD) improves LLM reasoning by using teacher-side privileged information (PI) to turn sparse verifier outcomes into dense token-level supervision. 
Existing methods usually assume trusted PI, such as reference answers or successful traces. 
We ask whether PI can instead come from an experience-derived skill bank, where retrieved skills are compact and reusable but may also be irrelevant or misleading. 
We propose Skill-Conditioned Gated Self-Distillation (\method{}), which formulates skill-based SD as teacher hypothesis validation rather than unconditional imitation. 
\method{} retrieves skill-mistake pairs, constructs a multi-teacher pool, and lets all skill-conditioned teachers score the same plain-prompt student rollout. 
The verifier validates each teacher's polarity: supporting a success or suppressing a failure gives positive supervision, while the opposite stance is reversed. 
A robust gated objective then distills informative teacher-student disagreements while suppressing uncertain or extreme signals. 
Experiments on multiple mathematical reasoning benchmarks show that \method{} consistently improves over GRPO and remains competitive with answer-conditioned OPSD under a weaker PI assumption. 
For example, on Qwen3-1.7B, \method{} outperforms GRPO by 6.2\% and OPSD by 1.7\% on AIME24, AIME25, and HMMT25.

\end{abstract}

\section{Introduction}

Reinforcement learning with verifiable rewards (RLVR) has become a central recipe for post-training large language models (LLMs) with reasoning capability. 
RLVR methods (e.g., GRPO \citep{deepseekmath}) sample solutions, verify the final answer, and improve the policy from environmental outcome feedback. 
This setting is attractive because it requires only a verifier, not token-level annotations or a larger teacher model. 
Its weakness is equally well known: the reward signal is sparse, delayed, and nearly uniform across the entire trajectory. 
On-policy distillation (OPD) \citep{agarwal2024policy} addresses this limitation by letting a stronger teacher provide dense token-level supervision on the student's sampled rollouts. 
On-policy self-distillation (SD)\footnote{To distinguish from another representative method OPSD, we uniformly use the abbreviation "SD" throughout this paper.} goes one step further: the same model acts as teacher and student under different contexts, removing the need for a separate teacher model \citep{zhao2026self,cui2026brief}. 
Therefore, SD promises the on-policy nature of RLVR together with richer credit assignment.

\begin{table*}[t]
\centering
\caption{\textbf{Comparison of representative post-training paradigms for LLM reasoning.} \method{} keeps on-policy dense self-supervision under a weaker PI assumption, using environment (verifier) outcomes and skill-conditioned teacher signals to decide the update direction.}
\label{tab:positioning}
\vspace{-1em}
\begingroup
\footnotesize
\setlength{\tabcolsep}{4pt}
\setlength{\aboverulesep}{0.35ex}
\setlength{\belowrulesep}{0.35ex}
\renewcommand{\arraystretch}{1.02}
\newcommand{\sgsdyes}{\textcolor{green!55!black}{$\checkmark$}}
\newcommand{\sgsdno}{\textcolor{red!75!black}{$\times$}}
\begin{tabular}{@{}>{\raggedright\arraybackslash}p{0.19\linewidth}
>{\centering\arraybackslash}p{0.13\linewidth}
>{\centering\arraybackslash}p{0.11\linewidth}
>{\centering\arraybackslash}p{0.12\linewidth}
>{\centering\arraybackslash}p{0.16\linewidth}
>{\raggedright\arraybackslash}p{0.20\linewidth}@{}}

\toprule
 & \textbf{Sampling} & \textbf{Signal} & \textbf{Teacher} & \textbf{PI Assumption} & \textbf{Direction Anchoring}
\\
\midrule
Distillation & \sgsdno{} off-policy & \sgsdyes{} dense & \sgsdno{} external & --- & \sgsdno{} teacher
\\
On-policy distillation & \sgsdyes{} on-policy & \sgsdyes{} dense & \sgsdno{} external & --- & \sgsdno{} teacher
\\
RLVR (e.g., GRPO) & \sgsdyes{} on-policy & \sgsdno{} sparse & --- & --- & \sgsdno{} environment
\\
OPSD & \sgsdyes{} on-policy & \sgsdyes{} dense & \sgsdyes{} self & \sgsdno{} strong & \sgsdno{} teacher
\\
\textbf{\method{}} (ours) & \sgsdyes{} on-policy & \sgsdyes{} dense & \sgsdyes{} self & \sgsdyes{} weak & \sgsdyes{} environment \& teacher
\\
\bottomrule
\end{tabular}
\endgroup
\vspace{-1em}
\end{table*}

The effectiveness of SD depends on the teacher-side privileged context information (PI). 
Prior work usually gives the teacher a reference solution, a successful trace, textual feedback, or demonstrations \citep{zhao2026self,yang2026selfdistilledrlvr,hubotter2026reinforcement,shenfeld2026selfdistillation}. 
In mathematical reasoning, this raises a natural question: can PI instead come from a structured \emph{skill bank}, as in prior work for agentic tasks \citep{wang2026skillsd}?
Skills \citep{anthropic3} are a common abstraction in agentic learning and memory-like systems: previous trajectories are compressed into reusable natural-language principles, tactics, and mistake warnings \citep{shinn2023reflexion,zhao2023expel,xia2026skillrl}. 
Compared with full traces or reference answers, skills are more compact and potentially more reusable across problems, making them appealing for mathematical reasoning in a weaker supervision regime.
Here, training may have only verifier feedback rather than a ground-truth answer.

However, skills introduce a harder reliability problem than the privileged contexts used in standard SD. 
A retrieved skill can be relevant, partially relevant, stale, or simply wrong for the current problem. 
More importantly, even a reasonable skill should not be copied unconditionally. 
For example: supporting a successful rollout is helpful, but supporting a failed rollout is harmful; suppressing an incorrect trajectory is useful, but suppressing a correct one is not. 
This leads to a central question: \emph{can experience-derived skills serve as training-only privileged information without assuming that every retrieved skill is trusted?}

Answering this question requires solving three challenges: 
\textbf{(1)} First, the privileged information is no longer a single, trusted source but a set of retrieved skill hypotheses with uncertain relevance. 
\textbf{(2)} Second, the value of a skill is outcome-dependent: the same teacher support can be beneficial or misleading depending on whether the sampled rollout succeeds. 
\textbf{(3)} Third, multiple skill-conditioned teachers can produce noisy or conflicting token-level signals, so the method must extract useful disagreement without being dominated by prompt artifacts or extreme logits. 

Therefore, we propose \textbf{Skill-Conditioned Gated Self-Distillation (\method{})}. 
\method{} maintains a structured bank of general skills and common mistake patterns distilled from past trajectories. 
For each problem, it retrieves relevant skills and mistakes, pairs them into a skill-conditioned multi-teacher pool, and lets all teachers score the same student rollout while the student itself sees only the plain problem. 
\method{} then combines each teacher's support with the rollout outcome to infer its polarity: a teacher is helpful when it supports a success or suppresses a failure, and harmful otherwise. 
A robust gated objective finally learns from informative teacher-student disagreements while downweighting ambiguous or extreme signals. 
These designs enable \method{} to achieve comparable performances even under a weaker PI assumption. 
Moreover, in contrast to prior distillation methods whose update direction is controlled by a teacher alone, \method{} derives the direction jointly from the environment outcome and the skill-conditioned teacher support. 
Tab. \ref{tab:positioning} summarizes the differences between \method{} and related paradigms.

Our contributions can be concluded as follows:
\begin{itemize}
\item We introduce skills and self-distillation to mathematical reasoning with verifier feedback, and formulate it as a teacher hypothesis validation problem rather than assuming retrieved skills should be imitated by default.
\item We propose \method{}, a multi-teacher, skill-conditioned self-distillation framework that infers teacher polarity from the alignment between teacher support and rollout outcome, then applies a robust gated objective.
\item We provide empirical results on several mathematical reasoning benchmarks, showing that our \method{} improves over GRPO and remains competitive with SD baselines. For example, on Qwen3-1.7B, \method{} improves the average score across AIME24, AIME25, and HMMT25 from 37.4 to 43.7.
\end{itemize}

\section{Our proposed \method{}}
\label{sec:method}

Let $x\sim\mathcal{D}$ denote a problem and $y$ a sampled reasoning trajectory. 
\method{} uses one policy model $\pi_\theta$ in two roles. 
The student is the plain-prompt policy $p_S(\cdot\mid x)=\pi_\theta(\cdot\mid x)$, which samples
\begin{equation}
y \sim p_S(\cdot \mid x).
\end{equation}
After generation, an answer verifier gives a scalar outcome $r\in\{-1,1\}$. 
The student is never given retrieved skills during rollout or evaluation. 
Skills are only injected as teacher-side PI during training.

\begin{figure*}[!h]
    \centering
    \includegraphics[width=0.95\linewidth]{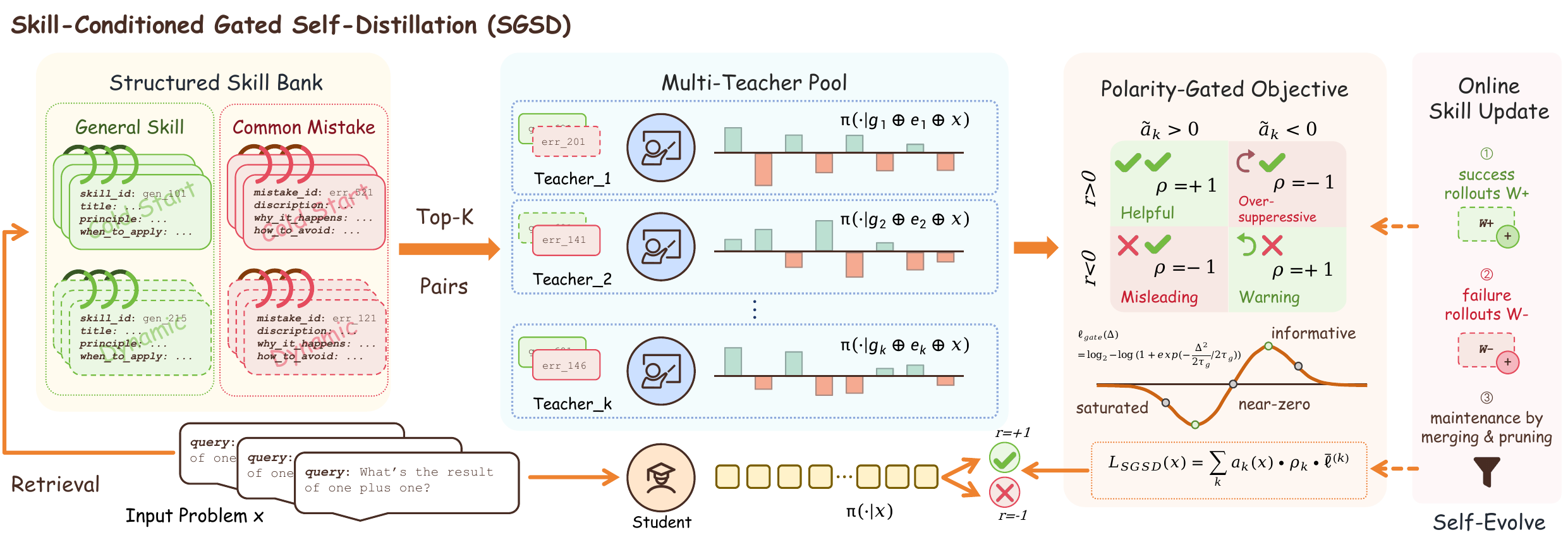}
    \caption{\textbf{Overview of \method{}.} The student samples a rollout from the plain problem, while retrieved skill--mistake pairs instantiate a pool of skill-conditioned teachers. Multiple teachers score the same rollout, the verifier outcome validates their polarity, and a gated distillation objective turns teacher-student gaps into dense supervision.}
    \label{fig:overview}
    \vspace{-0.4cm}
\end{figure*}

\subsection{Structured Skill Bank Construction}

\method{} uses a structured skill bank $\mathcal{B}=\{\mathcal{G},\mathcal{E}\}$ rather than raw historical trajectories\footnote{In this paper, a skill means an experience-derived natural-language principle, not an external tool or executable function.}. 
$\mathcal{G}$ contains \emph{general skills}, which encode reusable reasoning principles with a short title and conditions for use. 
$\mathcal{E}$ contains \emph{common mistake patterns}, which describe an error, why it occurs, and how to avoid it. 
Detailed schemas and prompt templates are given in App.~\ref{app:skill_bank} and App.~\ref{app:prompts}, respectively.

The bank is initialized before training by a cold-start extraction process. 
We first sample a subset of training problems as the cold-start seed set and generate solution traces for them with the initial policy. 
These traces are then converted into structured records. 
Specifically, successful traces are summarized into candidate general skills, failed traces are summarized into candidate mistake patterns, and redundant candidates are merged into a compact bank. 
This follows the same spirit as skill-library construction in agent learning \citep{xia2026skillrl}: preserve reusable experience while removing noisy trajectory details. 
App.~\ref{app:cold_start} gives the concrete construction procedure.

The central unit used by \method{} is a skill pair $(g_k,e_k)$, where $g_k\in\mathcal{G}$ provides positive guidance about what to do and $e_k\in\mathcal{E}$ provides negative guidance about what to avoid. 
Pairing the two creates a balanced teacher context instead of appending a large unstructured memory block. 
During training, each retrieved pair becomes a separate teacher-side PI block, and all resulting teachers score the same student rollout. 

\subsection{Skill-Conditioned Multi-Teacher Pool}
For a new problem $x$, \method{} performs semantic-similarity retrieval over the skill bank, obtaining:
\begin{equation}
\small
\mathcal{G}(x)=\{g_1,\ldots,g_{K_x}\},
\quad
\mathcal{E}(x)=\{e_1,\ldots,e_{K_x}\}.
\end{equation}
The $k$-th retrieved skill and mistake are concatenated with the problem to form the corresponding teacher-side PI:
\begin{equation}
c_k(x)=g_k\oplus e_k\oplus x,
\end{equation}
where $\oplus$ denotes prompt concatenation. 
These contexts define a pool of $K_x$ skill-conditioned teachers. The $k$-th teacher distribution is
\begin{equation}
p_T^{(k)}(y_t \mid x,y_{<t})
=
\pi_{\bar{\theta}}(y_t \mid c_k(x),y_{<t}),
\end{equation}
where $\bar{\theta}$ is the stop-gradient copy of the current policy parameters. 
Thus, all teachers share the same policy model as the student, but differ in the skill-conditioned PI they observe. 
The student distribution remains
\begin{equation}
p_S(y_t\mid x,y_{<t})=\pi_\theta(y_t\mid x,y_{<t}).
\end{equation}

\method{} assigns each teacher a relevance weight $\alpha_k(x)$ from the semantic retrieval scores. 
Let $s_k^{(g)}$ and $s_k^{(e)}$ be the similarity scores of the paired skill and mistake; we use the average score $s_k(x)=(s_k^{(g)}(x)+s_k^{(e)}(x))/2$ and normalize
\begin{equation}
\alpha_k(x)=
\frac{\exp(s_k(x))}
{\sum_{j=1}^{K_x}\exp(s_j(x))}.
\end{equation}

\subsection{Outcome-Validated Teacher Polarity}

The purpose of \method{} is not to imitate every teacher induced by retrieved skills, but to decide which teacher stances agree with the observed outcome and thus deserve to be distilled. 

For each generated token $y_t$, \method{} computes the teacher-student log-probability gap:
\begin{equation}
\small
\Delta_t^{(k)}
=
\log p_T^{(k)}(y_t\mid x,y_{<t})
-
\log p_S(y_t\mid x,y_{<t}).
\end{equation}
This gap is the token-level credit that appears in reverse-KL-style policy-gradient form. 
Positive $\Delta_t^{(k)}$ means teacher $k$ assigns higher probability to the sampled token than the student does and thus supports the token, while negative $\Delta_t^{(k)}$ means the teacher suppresses the sampled token. 

Next, define the plain teacher-level support score
\begin{equation}
a_k=
\frac{1}{T}\sum_{t=1}^{T}\Delta_t^{(k)}.
\vspace{-0.3cm}
\end{equation}
The direction of $a_k$ summarizes whether teacher $k$ supports or suppresses the sampled rollout over the whole trajectory. 
The outcome then validates this stance. Define
\begin{equation}
\small
\vspace{-0.1cm}
\operatorname{sgn}_{\mathrm{out}}(r)=
\begin{cases}
1, & r>0,\\
-1, & r<0,
\end{cases}
\vspace{-0.15cm}
\end{equation}
the plain polarity is
\begin{equation}
\rho_k^{\mathrm{plain}}
=
\operatorname{sgn}_{\mathrm{out}}(r)\operatorname{sgn}(a_k).
\end{equation}
Here, $\operatorname{sgn}(0)=0$. 
Therefore, a teacher is useful when it supports a successful rollout or suppresses a failed rollout, while it is misleading when it supports a failure or suppresses a success. 
Tab.~\ref{tab:polarity} summarizes the decision rule.

\begin{table}[t]
\centering
\caption{\textbf{Decision rule of teacher polarity.}
\method{} distills a skill-conditioned teacher when its polarity agrees with the verifier outcome, and reverses it otherwise.}
\label{tab:polarity}
\vspace{-0.8em}
\scriptsize
\setlength{\tabcolsep}{4pt}
\renewcommand{\arraystretch}{1.18}
\resizebox{\linewidth}{!}{
\begin{tabular}{@{}c c c c@{}}
\toprule
\textbf{Decision} 
& \textbf{Rule} 
& \textbf{Case} 
& \textbf{Interpretation} \\
\midrule
\multirow{2}{*}{\cellcolor{green!8}\textbf{Distill}}
& \multirow{2}{*}{$\mathrm{sign}(r)=\mathrm{sign}(a_k)$}
& $r>0,\ a_k>0$ 
& Helpful teacher \\
& 
& $r<0,\ a_k<0$ 
& Warning teacher \\
\midrule
\multirow{2}{*}{\cellcolor{orange!10}\textbf{Reverse}}
& \multirow{2}{*}{$\mathrm{sign}(r)\neq\mathrm{sign}(a_k)$}
& $r>0,\ a_k<0$ 
& Over-suppressive teacher \\
&
& $r<0,\ a_k>0$ 
& Misleading teacher \\
\bottomrule
\end{tabular}
}
\vspace{-0.8em}
\end{table}

\subsection{Robust Support Estimation}

The plain support score $a_k$ can be distorted by formatting tokens, prompt artifacts, or a few extreme token gaps. 
\method{} therefore estimates a robust support score $\widetilde{a}_k$ before deciding the final polarity. 
A token mask $m_t\in\{0,1\}$ first selects effective reasoning positions and removes special tokens, chat-template tokens, thinking delimiters, and pure-formatting tokens from aggregation. 
For support estimation, each gap is then clipped as
\begin{equation}
\widetilde{\Delta}_t^{(k)}
=
\operatorname{clip}(\Delta_t^{(k)}, -c_\Delta, c_\Delta),
\end{equation}
where $c_\Delta>0$ is the clipping bound. 
The robust support score is
\begin{equation}
\vspace{-0.2cm}
\widetilde{a}_k=
\frac{\sum_{t=1}^{T} m_t \widetilde{\Delta}_t^{(k)}}
{\sum_{t=1}^{T} m_t+\epsilon},
\end{equation}
where $\epsilon>0$ prevents division by zero. 
Furthermore, the final polarity uses a confidence threshold $\epsilon_a\ge0$ to allow uncertain teachers to be ignored:
\begin{equation}
\rho_k =
\begin{cases}
\operatorname{sgn}_{\mathrm{out}}(r)
\operatorname{sgn}(\widetilde{a}_k), & |\widetilde{a}_k|>\epsilon_a,\\
0, & |\widetilde{a}_k|\le \epsilon_a.
\end{cases}
\vspace{-0.2cm}
\end{equation}
Overall, masking preserves the effective reasoning tokens, clipping prevents outlier gaps from flipping the teacher-level stance, and thresholding creates a neutral zone for uncertain evidence. 

\subsection{Gated Distillation Objective}

After polarity decides whether a teacher should be distilled, reversed, or ignored for the current rollout, the remaining question is how strongly to learn from each teacher-student disagreement. 
Directly optimizing a sampled-token reverse-KL-style gap can overreact to very large discrepancies: extreme positive or negative gaps may dominate the optimization, while near-zero gaps carry little information. 
\method{} therefore uses a bounded gated loss:
\begin{equation}
\small
\ell_{\mathrm{gate}}(\Delta)
=
\log 2 - 
\log\left(1+\exp\left(-\frac{\Delta^2}{2\tau_g}\right)\right),
\vspace{-0.2cm}
\end{equation}
where $\tau_g>0$ controls the width of the informative region, and the $\log 2$ constant normalizes the loss to be zero at $\Delta=0$.  
The loss value is near zero when teacher and student agree, grows for moderate disagreement, and saturates for extreme mismatch.
And, the induced update remains bounded rather than being dominated by outlier gaps. 
Section~\ref{sec:theory} shows that its gradient yields a bounded policy-gradient-style correction and can be viewed as a sampled-token, polarity-aware approximation to local reverse-KL behavior. 

For each skill-conditioned teacher, the gated loss is
\begin{equation}
\bar{\ell}^{(k)}
=
\frac{\sum_{t=1}^{T} m_t \ell_{\mathrm{gate}}(\Delta_t^{(k)})}
{\sum_{t=1}^{T} m_t+\epsilon}.
\vspace{-0.1cm}
\end{equation}
The overall distillation objective is then given by:
\begin{equation}
\label{eq:loss-overall}
\mathcal{L}_{\mathrm{\method{}}}(x)
=
\sum_{k=1}^{K_x}
\alpha_k(x)\rho_k\bar{\ell}^{(k)}.
\vspace{-0.1cm}
\end{equation}
Since the teachers are stop-gradient, $\rho_k$ controls the direction of the update: helpful teachers distill their stance, misleading teachers reverse it, and uncertain teachers with $\rho_k=0$ contribute nothing.

\subsection{Online Skill Update and Maintenance}

\method{} can update its skill bank during training, so that retrieved skills can reflect the current policy's observed successes and failures. 
At fixed update intervals, recent student rollouts are collected into a window $\mathcal{W}$ with verifier outcomes. 
When recent success rate is already high, the update is skipped. 
Otherwise, the window is split into successful traces $\mathcal{W}^{+}$ and failed traces $\mathcal{W}^{-}$. 
Successful traces are summarized into new general-skill candidates, while failed traces are summarized into new common-mistake candidates. 
New candidates are then merged with existing bank entries, written back as dynamic entries, and pruned under a capacity limit to keep the bank compact.
App.~\ref{app:skill_evolution} gives the detailed update procedure, and Alg.~\ref{alg:procedure} summarizes the overall procedure of \method{}.

\begin{algorithm}[t]
\small
\caption{Overall Procedure of \method{}}
\label{alg:procedure}
\begin{algorithmic}[1]
\Require Training set $\mathcal{D}$, policy model $\pi_\theta$, skill bank $\mathcal{B}=\{\mathcal{G},\mathcal{E}\}$, retrieved pair count $K$
\For{each update}
\State Sample a minibatch $\mathcal{X}\subset\mathcal{D}$
\For{each problem $x\in\mathcal{X}$}
\State Retrieve $K$ ranked skill--mistake pairs $\{(g_k,e_k)\}_{k=1}^{K_x}$ from $\mathcal{B}$
\State Compute teacher weights $\{\alpha_k(x)\}_{k=1}^{K_x}$ from retrieval scores
\State Form skill-conditioned teacher contexts $c_k(x)=g_k\oplus e_k\oplus x$
\State Sample a plain-prompt student rollout $y\sim p_S(\cdot\mid x)$ and evaluate outcome $r$
\For{each teacher context $c_k(x)$}
\State Score the fixed rollout with the stop-gradient teacher: $p_T^{(k)}(y_t\mid x,y_{<t})=\pi_{\bar{\theta}}(y_t\mid c_k(x),y_{<t})$
\State Compute token gaps $\Delta_t^{(k)}$ and robust support $\widetilde{a}_k$
\State Infer polarity $\rho_k$ from $\widetilde{a}_k$ and $r$
\State Compute the gated teacher loss $\bar{\ell}^{(k)}$
\EndFor
\State Accumulate the per-example objective: $\mathcal{L}_{\mathrm{\method{}}}(x)=\sum_{k=1}^{K_x}\alpha_k(x)\rho_k\bar{\ell}^{(k)}$
\EndFor
\State Update $\theta$ with the minibatch \method{} objective
\State Update $\mathcal{B}$ from recent successful and failed rollouts
\EndFor
\end{algorithmic}
\end{algorithm}

\section{Theoretical Analysis}
\label{sec:theory}

This section analyzes why the gated distillation objective works. 
The key is its derivative
\begin{equation}
\small
g_{\tau}(\Delta)
=
\frac{\partial \ell_{\mathrm{gate}}(\Delta)}{\partial \Delta}
=
\frac{\Delta}
{\tau_g\left(1+\exp\left(\Delta^2/(2\tau_g)\right)\right)},
\end{equation}
because $g_{\tau}$ directly determines how each teacher-student gap enters the policy gradient. Detailed derivations are deferred to Apps.~\ref{app:proof} and~\ref{app:approx}, with a visualization in App.~\ref{app:gate_shape}.

\textbf{Policy-gradient form.}
Since teacher distributions are stop-gradient targets, we have
\begin{equation}
\nabla_\theta \Delta_t^{(k)}=
-\nabla_\theta \log p_S(y_t\mid x,y_{<t}).
\end{equation}
Let $Z=\sum_{t=1}^{T}m_t+\epsilon$. Substituting the derivative of $\ell_{\mathrm{gate}}$ into Eq.~\eqref{eq:loss-overall} gives
\begin{align}
\small
-\nabla_\theta \mathcal{L}_{\mathrm{\method{}}}(x)
&=
\sum_{k=1}^{K_x}\sum_{t=1}^{T}
\alpha_k(x)\rho_k
\frac{m_t}{Z}
g_{\tau}(\Delta_t^{(k)})
\nonumber\\
&\quad\cdot
\nabla_\theta \log p_S(y_t\mid x,y_{<t}).
\vspace{-0.3cm}
\end{align}
Thus \method{} is a policy-gradient-style update with dense token-level coefficient
\begin{equation}
W_t^{(k)}
=
\alpha_k(x)\rho_k
\frac{m_t}{Z}
g_{\tau}(\Delta_t^{(k)}).
\vspace{-0.3cm}
\end{equation}
In this sense, $W_t^{(k)}$ acts as an advantage-like dense credit.
The effectiveness of the loss reduces to whether $g_{\tau}$ induces the right gradient shape. 
Specifically, this gradient has three properties as follows.

\textbf{Sign consistency.}
We have
\begin{equation}
\operatorname{sign}(g_{\tau}(\Delta))=\operatorname{sign}(\Delta)
\quad (\Delta\neq0).
\end{equation}
Therefore, after applying $\rho_k$, helpful teachers reinforce their supporting tokens and misleading teachers reverse them.

\textbf{Bounded informative region.}
We introduce the following proposition:
\begin{proposition}
For every $\Delta\in\mathbb{R}$, we have
\begin{equation}
|g_{\tau}(\Delta)|\le \frac{1}{\sqrt{e\,\tau_g}},
\vspace{-0.15cm}
\end{equation}
Furthermore,
\begin{equation}
\small
g_{\tau}(\Delta)=\frac{\Delta}{2\tau_g}
+O\left(\frac{\Delta^3}{\tau_g^2}\right)
\quad \text{as }\Delta\to0,
\vspace{-0.15cm}
\end{equation}
while
\begin{equation}
\small
g_{\tau}(\Delta)=
\frac{\Delta}{\tau_g}
\exp\left(-\frac{\Delta^2}{2\tau_g}\right)
(1+o(1))
\quad \text{as }|\Delta|\to\infty .
\vspace{-0.15cm}
\end{equation}
\end{proposition}

This proposition captures the intended behavior of the gated loss: 
(1) Near zero, the update vanishes linearly, so negligible teacher-student disagreements do not inject noise. 
(2) For very large gaps, the exponential factor suppresses extreme mismatches instead of letting them dominate optimization. 
(3) Between these two regimes, the update is concentrated on medium gaps of order $\sqrt{\tau_g}$, which is exactly the informative region where the teacher disagrees enough to matter but not so much that the signal is likely driven by outliers. 

\textbf{Local reverse-KL approximation.}
The same derivative also explains why the gated loss behaves like a reasonable distillation objective in the informative region. When teacher and student are locally close, reverse KL admits the second-order expansion
\begin{equation}
D_{\mathrm{KL}}(p_T\,\|\,p_S)
=
\frac{1}{2}\sum_v p_{T,v}\Delta_v^2
+O(\|\Delta\|^3),
\end{equation}
where $\Delta_v=\log p_T(v\mid h_t)-\log p_S(v\mid h_t)$. 
Hence, the local reverse-KL gradient is linear in the log-probability gap. 
Our gated objective has the matching local form
\begin{equation}
g_{\tau}(\Delta)=\frac{\Delta}{2\tau_g}
+O\left(\frac{\Delta^3}{\tau_g^2}\right),
\end{equation}
so in the locally close regime, \method{} behaves like a sampled-token, polarity-aware reverse-KL correction up to a constant scale. 
Outside that regime, the gate deliberately departs from reverse KL by exponentially damping extreme gaps.

\section{Experiments}

\subsection{Experimental Setup}
\label{sec:exp_setup}

\textbf{Models and datasets.}
We use Qwen3-1.7B, Qwen3-4B, and Qwen3-8B as the base models \citep{yang2025qwen3}. 
We train on the English subset of DAPO-Math-17K \citep{yu2026dapo}, and evaluate out of domain on three challenging competition-style mathematical reasoning benchmarks: AIME24, AIME25, and HMMT25.

\textbf{Baselines and evaluation.}
We compare our method against \textbf{(1) GRPO} \citep{deepseekmath}, a standard RLVR baseline that optimizes group-relative rewards from sampled rollouts, and 
\textbf{(2) OPSD} \citep{zhao2026self}, a representative SD baseline that uses the reference solution as teacher-side PI for dense token-level self-supervision. 
Different from the standard OPSD implementation, we provide the teacher with the ground-truth answer rather than a full distillation trajectory. 
We also include several skill-augmented variants: 
\textbf{(1) Base+Skill} retrieves skills at inference; 
\textbf{(2) GRPO+Skill} injects retrieved skills into training rollouts but evaluates with the plain problem prompt; and 
\textbf{(3) OPSD+Skill} exposes both teacher and student to skills during training and also evaluates without skills. 
We train the base models for 200 steps, checkpoint every 25 steps, and report the highest avg@12 accuracy among recorded checkpoints.

\textbf{Implementation details.}
For each model, we randomly select 256 in-domain problems from the training set to build the cold-start skill bank. 
The teacher policy is synchronized with the student during training. 
Following OPSD, we disable thinking mode for the student and enable thinking mode for the teacher. 
We use Qwen3-Embedding-0.6B as the retriever to retrieve the top-8 skill pairs, and compute full-vocabulary distillation. 
The skill bank is updated every 25 steps. 
All experiments are conducted on eight NVIDIA A800 GPUs with LoRA adaptation \citep{hu2022lora}, TRL framework \citep{vonwerra2020trl}.
For more experimental details, please refer to App.~\ref{app:exp_details}.

\subsection{Main Results}

\begin{table*}[t]
\centering
\caption{\textbf{Main comparison results on mathematical reasoning benchmarks.} The best and second-best results are highlighted in \textbf{bold} and \underline{underlined}, respectively. We report the avg@12 accuracy for the best checkpoint on AIME24, AIME25, and HMMT25. Skill-conditioned \method{} improves over GRPO, and remains competitive with answer-conditioned OPSD under a weaker PI assumption.}
\label{tab:main}
\vspace{-1em}
\small
\begin{tabular}{llcccc}
\toprule
Model & Method & AIME24 & AIME25 & HMMT25 & Avg. \\
\midrule
\multirow{7}{*}{Qwen3-1.7B}
& Base & 51.1 & 36.9 & 24.2 & 37.4 \\
& Base+Skill & 49.2 & 38.3 & 20.3 & 35.9 \\
& GRPO & 50.0 & 36.7 & 25.8 & 37.5 \\
& GRPO+Skill & 52.2 & 38.9 & 25.8 & 39.0 \\
& OPSD & \underline{55.8} & \underline{43.3} & 26.9 & \underline{42.0} \\
& OPSD+Skill & 54.4 & 40.3 & \underline{27.2} & 40.6 \\
\rowcolor{bggray} & \method{} & \textbf{57.8} & \textbf{43.6} & \textbf{29.7} & \textbf{43.7} \\
\midrule
\multirow{7}{*}{Qwen3-4B}
& Base & 73.9 & \underline{69.7} & 43.3 & 62.3 \\
& Base+Skill & 71.9 & 65.6 & 42.2 & 59.9 \\
& GRPO & \textbf{76.1} & 66.7 & 45.3 & 62.7 \\
& GRPO+Skill & 73.9 & 65.3 & 44.7 & 61.3 \\
& OPSD & \underline{75.3} & 69.2 & \underline{46.1} & \underline{63.5} \\
& OPSD+Skill & \underline{75.3} & 66.7 & 44.7 & 62.2 \\
\rowcolor{bggray} & \method{} & \underline{75.3} & \textbf{70.8} & \textbf{46.7} & \textbf{64.3} \\
\midrule
\multirow{7}{*}{Qwen3-8B}
& Base & 75.3 & 66.1 & 45.0 & 62.1 \\
& Base+Skill & 78.3 & 68.1 & 40.8 & 62.4 \\
& GRPO & \textbf{79.2} & 69.7 & 46.1 & 65.0 \\
& GRPO+Skill & 78.3 & 66.9 & 45.0 & 63.4 \\
& OPSD & \textbf{79.2} & \textbf{73.1} & \textbf{48.1} & \textbf{66.8} \\
& OPSD+Skill & 77.8 & 65.3 & 43.9 & 62.3 \\
\rowcolor{bggray} & \method{} & \underline{78.9} & \underline{70.6} & \underline{46.9} & \underline{65.5} \\
\bottomrule
\end{tabular}
\end{table*}

\begin{figure*}[!h]
    \centering
    \includegraphics[width=0.9\linewidth]{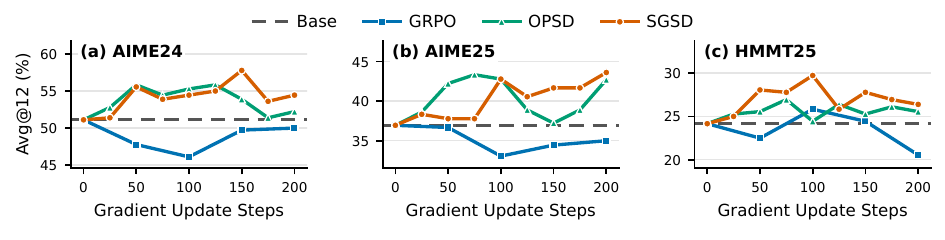}
    \vspace{-0.5cm}
    \caption{\textbf{Training dynamics on Qwen3-1.7B.} GRPO shows limited improvement under sparse outcome rewards, while \method{} maintains more stable gains than OPSD in the later training stage.}
    \label{fig:dynamics}
\end{figure*}

\textbf{Skill-conditioned SD improves reasoning under weaker PI.}
Tab.~\ref{tab:main} shows the main comparison.
GRPO brings only limited gains, which is consistent with the sparsity of final-answer, trajectory-level rewards. 
OPSD and \method{} turn sparse rewards into dense token-level supervision, yielding stronger improvements across model scales. 
Surprisingly, \method{} performs competitively with OPSD, and even achieves better performance on smaller models.
For example, on Qwen3-1.7B, \method{} improves the average score from 37.4\% to 43.7\%, outperforming GRPO by 6.2\% and OPSD by 1.7\%. 
This is achieved under a weaker PI assumption: OPSD uses a trusted reference answer as PI, whereas \method{} only assumes a verifier and a skill pair that may be misleading. 
Although \method{} still underperforms OPSD on Qwen3-8B, the results already suggest that outcome-validated skill-conditioned teachers can also approach answer-conditioned teachers for reasoning.

\textbf{Skill injection alone is not sufficient.}
The skill-augmented variants do not perform well in Tab.~\ref{tab:main}, showing that the gains do not come from simply exposing the model to more skills. 
Base+Skill can underperform the plain base model, likely because a single retrieval step does not always return skills that are useful for the current reasoning problem. 
GRPO+Skill and OPSD+Skill show a similar issue: when skills are directly visible during training, the model may rely on them to solve the problem, rather than internalizing them to reason without skills. 
This issue is even clearer for OPSD+Skill, where teacher and student observe the same skills, leaving little teacher-student contrast for distillation. 
\method{} instead keeps skills on the teacher side only, using a multi-teacher pool so that skill-conditioned teachers can provide relevant guidance for the student to internalize.

\subsection{Training Dynamics}
Fig.~\ref{fig:dynamics} illustrates the training dynamics of on Qwen3-1.7B over 200 update steps. 
As shown, GRPO often stays below the base model, reflecting the weak learning signal from sparse outcome rewards. 
OPSD improves rapidly in the early stage, but gradually degrades after the mid-training checkpoints. 
In contrast, \method{} shows a more stable improvement pattern, reaching 42.3\% average performance at step 100 and keeps a clear advantage over OPSD in the later stage. 
By validating teacher support with verifier outcomes and suppressing uncertain or extreme updates, \method{} keeps useful dense supervision available as both the policy and the skill bank evolve online.

\subsection{Ablation Studies}
\begin{figure*}[!h]
    \centering
    \includegraphics[width=0.99\linewidth]{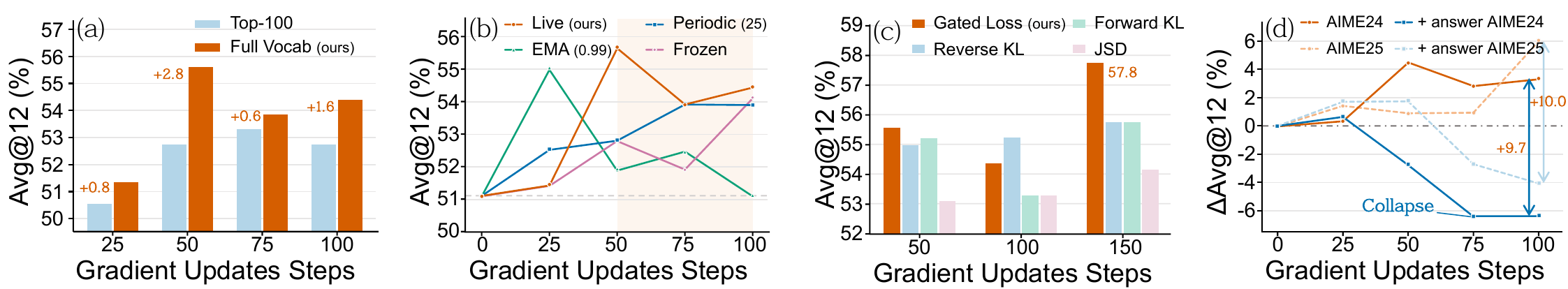}
    \caption{\textbf{Further discussions on Qwen3-1.7B.} (a) Full-vocabulary distillation consistently outperforms Top-100 support distillation on AIME24. (b) The live (synchronized) teacher used by \method{} reaches the strongest later checkpoints, while alternative update strategies either lag or fluctuate. (c) The gated objective obtains the highest peak among divergence-style losses on AIME24. (d) Compared with vanilla \method{}, adding OPSD-style answer PI to skill-conditioned teachers causes a large performance drop and thus collapse on both AIME24 and AIME25.}
    \label{fig:discussion}
\end{figure*}

We ablate the main components of \method{} on Qwen3-1.7B model, AIME24 dataset.

\textbf{Robust support estimation improves training stability.}
As shown in Tab.~\ref{tab:ablation-robust}, removing either token masking, clipping, or thresholding weakens model performance, and this phenomenon becomes more obvious in the later training stage.
Removing all three even falls below the base model at step 100.
Together, these components enable robust and stable distillation.

\begin{figure}[t]
    \centering
    \vspace{-0.2cm}
    \includegraphics[width=0.7\linewidth]{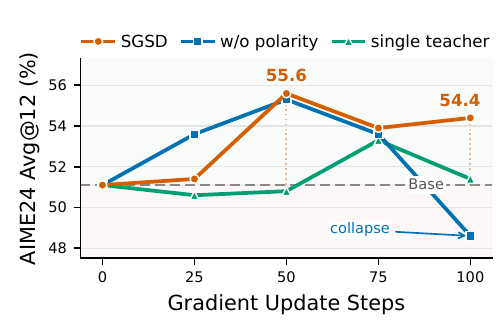}
    \vspace{-0.3cm}
    \caption{\textbf{Core ablations on Qwen3-1.7B.} Removing polarity causes late-stage collapse, while replacing the teacher pool with a single teacher remains below \method{}.}
    \vspace{-0.5cm}
    \label{fig:ablation}
\end{figure}

\textbf{Teacher polarity and diversity are essential.}
Fig.~\ref{fig:ablation} shows that removing polarity yields a short-term gain but eventually leads to collapse. 
Without polarity, all teachers are distilled in the same direction, introducing misleading supervision. 
Moreover, the variant without the multi-teacher pool consistently underperforms \method{}. 
This highlights the importance of teacher diversity: a single retrieved skill--mistake pair may be irrelevant or only partially useful for the problem, while the multi-teacher pool provides multiple hypotheses and a better chance to distill relevant guidance.
\begin{table}[t]
\centering
\caption{\textbf{Robust support estimation ablations on Qwen3-1.7B, AIME24.} Arrows indicate changes relative to the Avg@12 accuracy of base model.}
\vspace{-1em}
\label{tab:ablation-robust}
\small
\resizebox{0.75\linewidth}{!}{
\begin{tabular}{@{}lcc@{}}
\toprule
\textbf{Method} & \textbf{Step 50} & \textbf{Step 100} \\
\midrule
\textbf{Base}
& \multicolumn{2}{c}{51.1} \\
\rowcolor{bggray} \textbf{\method{}}
& \textbf{55.6} {\scriptsize\textcolor{green!55!black}{$\uparrow$4.5}}
& \textbf{54.4} {\scriptsize\textcolor{green!55!black}{$\uparrow$3.3}} \\
\hspace{2mm}w/o token mask
& 53.1 {\scriptsize\textcolor{green!55!black}{$\uparrow$2.0}}
& 54.2 {\scriptsize\textcolor{green!55!black}{$\uparrow$3.1}} \\
\hspace{2mm}w/o clipping
& 53.6 {\scriptsize\textcolor{green!55!black}{$\uparrow$2.5}}
& 52.5 {\scriptsize\textcolor{green!55!black}{$\uparrow$1.4}} \\
\hspace{2mm}w/o threshold
& 54.4 {\scriptsize\textcolor{green!55!black}{$\uparrow$3.3}}
& 51.4 {\scriptsize\textcolor{green!55!black}{$\uparrow$0.3}} \\
\hspace{2mm}w/o all three
& 54.2 {\scriptsize\textcolor{green!55!black}{$\uparrow$3.1}}
& 48.1 {\scriptsize\textcolor{red}{$\downarrow$3.0}} \\
\bottomrule
\end{tabular}
}
\end{table}

\subsection{Further Discussions}

We further study several design choices to understand when skill-conditioned SD remains reliable.

\textbf{Full-vocabulary vs. Top-K support.}
Fig.~\ref{fig:discussion}(a) compares the default full-vocabulary distillation\footnote{Note that the objective remains defined on the sampled tokens through gap $\Delta_t^{(k)}$. The term "full-vocabulary" indicates that $p_T(\cdot)$ and $p_S(\cdot)$ are normalized over the full vocabulary and then used to compute $\Delta_t^{(k)}$ and support score $a_k$.} with a top-100 support approximation.
For top-$k$ support, we re-normalize both distributions on the teacher's local top-$k$ support along with the sampled token, compute the resulting truncated local reverse-KL term \citep{fu2026revisiting}, and apply the same gated objective.
As shown, full-vocabulary distillation performs better at every checkpoint.
This suggests that tail probabilities still help calibrate teacher-student log-probability gaps, and truncating to top-$k$ can distort support estimation and weaken polarity-aware supervision.

\textbf{Teacher policy update.}
Fig.~\ref{fig:discussion}(b) studies alternative teacher policy update strategies, including frozen, periodic, and EMA teachers. 
Synchronizing the teacher with the student is an aggressive choice, since both policies may confidently drift together during self-distillation.
Nevertheless, the live teacher used by \method{} surprisingly performs well, reaching the highest peak accuracy and remain stable. 
This suggests that outcome-derived polarity and robust support estimation can stabilize training and make fresh teacher signals usable.

\textbf{Divergence-style distillation objective.}
Fig.~\ref{fig:discussion}(c) compares our gated loss with several standard divergence-style objectives.
Reverse KL is stable and competitive, but the gated loss obtains the highest peak, while forward KL and JSD are generally weaker.
This matches the intended role of the gate: it preserves local reverse-KL-style correction for informative gaps while suppressing negligible disagreements and extreme mismatches.

\textbf{Adding answer PI to \method{}.}
Fig.~\ref{fig:discussion}(d) adds the OPSD-style answer PI to the skill-conditioned teacher context. 
We observe that although this variant slightly improves at early stage, it soon consistently degrades and finally collapses on both AIME24 and AIME25, leaving \method{} ahead by 9.7\% and 10.0\% at step 100.
One possible reason is that answer PI and polarity-controlled skill PI impose different information structures: when some teachers are intentionally reversed, attaching the same answer to every teacher may make the direction signal inconsistent.
Thus, \method{} benefits from reusable skill PI, rather than simply giving the teacher richer privileged information.

\section{Related Work}

\textbf{Outcome-driven reasoning optimization.}
Large language model reasoning is commonly improved with preference, reward, or verifiable outcome signals. 
RLHF \citep{ouyang2022training} and DPO \citep{rafailov2023direct} optimize behavior from human or preference supervision, while RLVR methods \citep{rlvr-emp,ma2026thinking, liu2026automated} optimize answers that can be checked automatically, with GRPO-style training \citep{deepseekmath, zhai2026rewards} widely used for mathematical reasoning. 
Recent work further studies how to make outcome supervision more informative through self-rewarding signals and turn-level credit assignment \citep{yuan2024selfrewarding,hu2026selfinducedoutcomepotentialturnlevel}, inference-time budget control, decoding-time task adaptation, retrieval rewards, and noise-tolerant training schedules \citep{fang2026inferencetimebudgetcontrolllm,hu2026distributionaligned,hu2026optimizingagenticreasoningretrieval,xu2026explorationexploitationtwostageentropy}.
These methods face the same limitation: rewards are sparse, delayed and nearly uniform over the trajectories. 

\textbf{On-policy distillation and self-distillation.}
Knowledge distillation transfers behavior from a teacher to a student \citep{hinton2015distilling}. 
Traditional distillation is usually performed on a fixed dataset, which creates a mismatch between the trajectories seen during training and those produced at inference time. 
On-policy distillation (OPD) \citep{agarwal2024policy} addresses this issue by letting the student sample its own trajectories while a stronger teacher provides dense token-level supervision on those trajectories. 
For reasoning models, this makes distillation more compatible with post-training settings that need fine-grained credit assignment. 
Recent on-policy self-distillation (SD) methods go one step further by removing the external teacher: the same model acts as both teacher and student, while the teacher is conditioned on privileged context such as reference solutions, revised attempts, textual feedback, or context-conditioned guidance \citep{zhao2026self,hubotter2026reinforcement,shenfeld2026selfdistillation,ye2026policy,chen2025test}. 
Other extensions study self-revision, unified distillation objectives, or sample routing under the same goal of converting sparse outcomes into dense token-level updates \citep{he2026selfdistillationzero,jin2026unisd,li2026samplerouting}. 

\textbf{Experience, skills, and agent memory.}
A complementary line of work represents prior behavior as reusable natural-language experience. 
Chain-of-thought (CoT) \citep{wei2022chainofthought} and STaR \citep{zelikman2024star} reuse reasoning traces for self-improvement. 
Reflexion, Voyager, ExpeL \citep{shinn2023reflexion,wang2023voyager,zhao2023expel}, and collaborative LLM-agent systems \citep{hu2024agentscodriverlargelanguagemodel,10976336,zhu2026unified} study how language-model agents reuse experience or coordinate decisions across tasks.
Recent skill-based methods make this abstraction more explicit by distilling trajectories into reusable skills \citep{ni2026trace2skill} or evolving skill libraries over time \citep{wang2025reinforcement,zhang2026memskill,xia2026skillrl,zhang2026memmark,webcogreasoner}. 
\method{} does not inject retrieved skills into the student prompt at evaluation. 
Instead, skills are used as uncertain teacher-side hypotheses during training.

\nocite{ma2025instruct, liu2026all, li2026correcting, su2026sell, zhang2026temporal, shi2026androtmem, liu2026toolanchor, a2eval, huang2026gui}

\section{Conclusion}

We introduced \method{}, a skill-conditioned self-distillation framework for mathematical reasoning with verifier feedback. 
Rather than assuming that retrieved skills should always be trusted, \method{} treats skill--mistake pairs as teacher hypotheses, validates their polarity with rollout outcomes, and learns from informative teacher-student disagreements through a robust gated objective. 
Experiments on competition-style mathematical reasoning benchmarks show that \method{} improves over GRPO and simple skill-injection baselines, while remaining competitive with answer-conditioned OPSD under a weaker PI assumption. 
These results suggest that experience-derived skills can provide useful dense supervision when their direction is validated by on-policy evidence.

\section*{Limitations}
\method{} is evaluated mainly on mathematical reasoning tasks with automatic answer verification. 
This setting is well suited to our goal of studying outcome-validated skill-conditioned teachers, while extensions to open-ended generation may require a calibrated judge or another reliable outcome signal. 
Our skill bank uses a simple cold-start construction and rank-wise pairing between general skills and mistake patterns; more adaptive retrieval, pairing, or weighting strategies may further improve the teacher pool. 
The main experiments report the best checkpoint among recorded training steps for comparability with recent self-distillation work, so future studies could complement this protocol with held-out model selection. 
Finally, most ablations are conducted on Qwen3-1.7B to control computational cost, leaving a fuller sweep across larger models and additional hyperparameters for future work.

\section*{Ethical Considerations}
This work studies post-training for mathematical reasoning using public-style benchmark problems, automatically generated model traces, and verifiable answers. 
It does not involve human subjects, private user data, or sensitive attributes. 
Because \method{} stores experience-derived skills, care should be taken to keep the skill bank aligned with the intended training data and to avoid mixing evaluation problems into skill construction. 
The method also uses verifier outcomes to decide teacher polarity, so applications beyond mathematical reasoning should ensure that the outcome signal is appropriate for the task. 
As with other reasoning post-training methods, improved mathematical performance should not be interpreted as broad reliability in high-stakes domains without separate validation.

\bibliography{custom}

@article{zhao2026self,
  title={{S}elf-{D}istilled {R}easoner: {O}n-{P}olicy {S}elf-{D}istillation for {L}arge {L}anguage {M}odels},
  author={Zhao, Siyan and Xie, Zhihui and Liu, Mengchen and Huang, Jing and Pang, Guan and Chen, Feiyu and Grover, Aditya},
  journal={arXiv preprint},
  year={2026}
}

@article{hubotter2026reinforcement,
  title={{R}einforcement {L}earning via {S}elf-{D}istillation},
  author={H{\"u}botter, Jonas and L{\"u}beck, Frederike and Behric, Lejs and Baumann, Anton and Bagatella, Marco and Marta, Daniel and Hakimi, Ido and Shenfeld, Idan and Buening, Thomas Kleine and Guestrin, Carlos and others},
  journal={arXiv preprint},
  year={2026}
}

@article{ye2026policy,
  title={{O}n-{P}olicy {C}ontext {D}istillation for {L}anguage {M}odels},
  author={Ye, Tianzhu and Dong, Li and Wu, Xun and Huang, Shaohan and Wei, Furu},
  journal={arXiv preprint},
  year={2026}
}

@article{yang2026selfdistilledrlvr,
  title={{S}elf-{D}istilled {RLVR}},
  author={Yang, Chenxu and Qin, Chuanyu and Si, Qingyi and Chen, Minghui and Gu, Naibin and Yao, Dingyu and Lin, Zheng and Wang, Weiping and Wang, Jiaqi and Duan, Nan},
  journal={arXiv preprint},
  year={2026}
}

@article{he2026selfdistillationzero,
  title={{S}elf-{D}istillation {Z}ero: {S}elf-{R}evision {T}urns {B}inary {R}ewards into {D}ense {S}upervision},
  author={He, Yinghui and Kaur, Simran and Bhaskar, Adithya and Yang, Yongjin and Liu, Jiarui and Ri, Narutatsu and Fowl, Liam and Panigrahi, Abhishek and Chen, Danqi and Arora, Sanjeev},
  journal={arXiv preprint},
  year={2026}
}

@article{jin2026unisd,
  title={{U}ni{S}{D}: {T}owards a {U}nified {S}elf-{D}istillation {F}ramework for {L}arge {L}anguage {M}odels},
  author={Jin, Yiqiao and Wang, Yiyang and Fu, Lucheng and Xiao, Yijia and Luo, Yinyi and Liu, Haoxin and Prakash, B Aditya and Hester, Josiah and Wang, Jindong and Kumar, Srijan},
  journal={arXiv preprint},
  year={2026}
}

@article{li2026samplerouting,
  title={{U}nifying {G}roup-{R}elative and {S}elf-{D}istillation {P}olicy {O}ptimization via {S}ample {R}outing},
  author={Li, Gengsheng and Yang, Tianyu and Fang, Junfeng and Song, Mingyang and Zheng, Mao and Guo, Haiyun and Zhang, Dan and Wang, Jinqiao and Chua, Tat-Seng},
  journal={arXiv preprint},
  year={2026}
}

@article{shenfeld2026selfdistillation,
  title={{S}elf-{D}istillation {E}nables {C}ontinual {L}earning},
  author={Shenfeld, Idan and Damani, Mehul and H{\"u}botter, Jonas and Agrawal, Pulkit},
  journal={arXiv preprint},
  year={2026}
}

@article{deepseekmath,
  title={{D}eepseekmath: {P}ushing the {L}imits of {M}athematical {R}easoning in {O}pen {L}anguage {M}odels},
  author={Shao, Zhihong and Wang, Peiyi and Zhu, Qihao and Xu, Runxin and Song, Junxiao and Bi, Xiao and Zhang, Haowei and Zhang, Mingchuan and Li, YK and Wu, Yang and others},
  journal={arXiv preprint},
  year={2024}
}

@article{rafailov2023direct,
  title={{D}irect {P}reference {O}ptimization: {Y}our {L}anguage {M}odel {I}s {S}ecretly a {R}eward {M}odel},
  author={Rafailov, Rafael and Sharma, Archit and Mitchell, Eric and Manning, Christopher D and Ermon, Stefano and Finn, Chelsea},
  journal={Advances in neural information processing systems},
  volume={36},
  pages={53728--53741},
  year={2023}
}

@article{yang2025qwen3,
  title={{Q}wen3 {T}echnical {R}eport},
  author={Yang, An and Li, Anfeng and Yang, Baosong and Zhang, Beichen and Hui, Binyuan and Zheng, Bo and Yu, Bowen and Gao, Chang and Huang, Chengen and Lv, Chenxu and others},
  journal={arXiv preprint},
  year={2025}
}

@article{wei2022chainofthought,
  title={{C}hain-of-{T}hought {P}rompting {E}licits {R}easoning in {L}arge {L}anguage {M}odels},
  author={Wei, Jason and Wang, Xuezhi and Schuurmans, Dale and Bosma, Maarten and Xia, Fei and Chi, Ed and Le, Quoc V and Zhou, Denny and others},
  journal={Advances in neural information processing systems},
  volume={35},
  pages={24824--24837},
  year={2022}
}

@article{zelikman2024star,
  title={{S}tar: {B}ootstrapping {R}easoning with {R}easoning},
  author={Zelikman, Eric and Wu, Yuhuai and Mu, Jesse and Goodman, Noah},
  journal={Advances in Neural Information Processing Systems},
  volume={35},
  pages={15476--15488},
  year={2022}
}

@article{yuan2024selfrewarding,
  title={{S}elf-{R}ewarding {L}anguage {M}odels},
  author={Yuan, Weizhe and Pang, Richard Yuanzhe and Cho, Kyunghyun and Li, Xian and Sukhbaatar, Sainbayar and Xu, Jing and Weston, Jason},
  journal={arXiv preprint},
  year={2024}
}

@article{ouyang2022training,
  title={{T}raining {L}anguage {M}odels to {F}ollow {I}nstructions with {H}uman {F}eedback},
  author={Ouyang, Long and Wu, Jeffrey and Jiang, Xu and Almeida, Diogo and Wainwright, Carroll and Mishkin, Pamela and Zhang, Chong and Agarwal, Sandhini and Slama, Katarina and Ray, Alex and others},
  journal={Advances in neural information processing systems},
  volume={35},
  pages={27730--27744},
  year={2022}
}

@article{shinn2023reflexion,
  title={{R}eflexion: {L}anguage {A}gents with {V}erbal {R}einforcement {L}earning},
  author={Shinn, Noah and Cassano, Federico and Gopinath, Ashwin and Narasimhan, Karthik and Yao, Shunyu},
  journal={Advances in neural information processing systems},
  volume={36},
  pages={8634--8652},
  year={2023}
}

@article{wang2023voyager,
  title={{V}oyager: {A}n {O}pen-{E}nded {E}mbodied {A}gent with {L}arge {L}anguage {M}odels},
  author={Wang, Guanzhi and Xie, Yuqi and Jiang, Yunfan and Mandlekar, Ajay and Xiao, Chaowei and Zhu, Yuke and Fan, Linxi and Anandkumar, Anima},
  journal={arXiv preprint},
  year={2023}
}

@inproceedings{zhao2023expel,
  title={{E}xpel: {L}lm {A}gents {A}re {E}xperiential {L}earners},
  author={Zhao, Andrew and Huang, Daniel and Xu, Quentin and Lin, Matthieu and Liu, Yong-Jin and Huang, Gao},
  booktitle={Proceedings of the AAAI Conference on Artificial Intelligence},
  volume={38},
  number={17},
  pages={19632--19642},
  year={2024}
}

@article{hu2026selfinducedoutcomepotentialturnlevel,
  title={{S}elf-{I}nduced {O}utcome {P}otential: {T}urn-{L}evel {C}redit {A}ssignment for {A}gents {W}ithout {V}erifiers},
  author={Hu, Senkang and Dai, Yong and Han, Xudong and Fang, Zhengru and Zhao, Yuzhi and Kwong, Sam Tak Wu and Fang, Yuguang},
  journal={arXiv preprint},
  year={2026}
}

@article{fang2026inferencetimebudgetcontrolllm,
  title={{I}nference-{T}ime {B}udget {C}ontrol for {L}{L}{M} {S}earch {A}gents},
  author={Fang, Zhengru and Hu, Senkang Forest and Chang, Zhonghao and Guo, Yu and Tao, Yihang and Liu, Hongyao and Ruan, Mengzhe and Huang, Jun and Fang, Yuguang},
  journal={arXiv preprint},
  year={2026}
}

@article{hu2026optimizingagenticreasoningretrieval,
  title={{O}ptimizing {A}gentic {R}easoning with {R}etrieval via {S}ynthetic {S}emantic {I}nformation {G}ain {R}eward},
  author={Hu, Senkang and Dai, Yong and Zhao, Yuzhi and Tao, Yihang and Guo, Yu and Fang, Zhengru and Kwong, Sam Tak Wu and Fang, Yuguang},
  journal={arXiv preprint},
  year={2026}
}

@article{xu2026explorationexploitationtwostageentropy,
  title={{F}rom {E}xploration to {E}xploitation: {A} {T}wo-{S}tage {E}ntropy {R}{L}{V}{R} {A}pproach for {N}oise-{T}olerant {M}{L}{L}{M} {T}raining},
  author={Xu, Donglai and Yang, Hongzheng and Zhao, Yuzhi and Zhang, Pingping and Chen, Jinpeng and Ma, Wenao and Hou, Zhijian and Wu, Mengyang and Li, Xiaolei and Hu, Senkang and others},
  journal={arXiv preprint},
  year={2025}
}

@article{hinton2015distilling,
  title={{D}istilling the {K}nowledge in a {N}eural {N}etwork},
  author={Hinton, Geoffrey and Vinyals, Oriol and Dean, Jeff},
  journal={arXiv preprint},
  year={2015}
}

@article{xia2026skillrl,
  title={{S}killrl: {E}volving {A}gents via {R}ecursive {S}kill-{A}ugmented {R}einforcement {L}earning},
  author={Xia, Peng and Chen, Jianwen and Wang, Hanyang and Liu, Jiaqi and Zeng, Kaide and Wang, Yu and Han, Siwei and Zhou, Yiyang and Zhao, Xujiang and Chen, Haifeng and others},
  journal={arXiv preprint},
  year={2026}
}

@article{wang2026skillsd,
  title={{S}kill-{S}d: {S}kill-{C}onditioned {S}elf-{D}istillation for {M}ulti-{T}urn {L}lm {A}gents},
  author={Wang, Hao and Wang, Guozhi and Xiao, Han and Zhou, Yufeng and Pan, Yue and Wang, Jichao and Xu, Ke and Wen, Yafei and Ruan, Xiaohu and Chen, Xiaoxin and others},
  journal={arXiv preprint},
  year={2026}
}

@article{ni2026trace2skill,
  title={{T}race2skill: {D}istill {T}rajectory-{L}ocal {L}essons into {T}ransferable {A}gent {S}kills},
  author={Ni, Jingwei and Liu, Yihao and Liu, Xinpeng and Sun, Yutao and Zhou, Mengyu and Cheng, Pengyu and Wang, Dexin and Zhao, Erchao and Jiang, Xiaoxi and Jiang, Guanjun},
  journal={arXiv preprint},
  year={2026}
}

@article{cui2026brief,
  title={{A} {B}rief {O}verview: {O}n-{P}olicy {S}elf-{D}istillation {I}n {L}arge {L}anguage {M}odels},
  author={Cui, Fangming and Li, Sunan and Li, Jiahong},
  journal={arXiv preprint},
  year={2026}
}

@inproceedings{agarwal2024policy,
  title={{O}n-{P}olicy {D}istillation of {L}anguage {M}odels: {L}earning from {S}elf-{G}enerated {M}istakes},
  author={Agarwal, Rishabh and Vieillard, Nino and Zhou, Yongchao and Stanczyk, Piotr and Ramos Garea, Sabela and Geist, Matthieu and Bachem, Olivier},
  booktitle={International Conference on Learning Representations},
  volume={2024},
  pages={21246--21263},
  year={2024}
}

@misc{anthropic3,
title={{T}he {C}laude 3 {M}odel {F}amily: {O}pus, {S}onnet, {H}aiku},
year={2024},
author={Anthropic}
}

@article{zhang2026memskill,
  title={{M}em{S}kill: {L}earning and {E}volving {M}emory {S}kills for {S}elf-{E}volving {A}gents},
  author={Zhang, Haozhen and Long, Quanyu and Bao, Jianzhu and Feng, Tao and Zhang, Weizhi and Yue, Haodong and Wang, Wenya},
  journal={arXiv preprint},
  year={2026}
}

@article{wang2025reinforcement,
  title={{R}einforcement {L}earning for {S}elf-{I}mproving {A}gent with {S}kill {L}ibrary},
  author={Wang, Jiongxiao and Yan, Qiaojing and Wang, Yawei and Tian, Yijun and Mishra, Soumya Smruti and Xu, Zhichao and Gandhi, Megha and Xu, Panpan and Cheong, Lin Lee},
  journal={arXiv preprint},
  year={2025}
}

@article{yu2026dapo,
  title={{D}apo: {A}n {O}pen-{S}ource {L}lm {R}einforcement {L}earning {S}ystem at {S}cale},
  author={Yu, Qiying and Zhang, Zheng and Zhu, Ruofei and Yuan, Yufeng and Zuo, Xiaochen and Yue, Yu and Dai, Weinan and Fan, Tiantian and Liu, Gaohong and Liu, Lingjun and others},
  journal={Advances in Neural Information Processing Systems},
  volume={38},
  pages={113222--113244},
  year={2026}
}

@article{hu2022lora,
  title={{L}ora: {L}ow-{R}ank {A}daptation of {L}arge {L}anguage {M}odels},
  author={Hu, Edward J and Shen, Yelong and Wallis, Phillip and Allen-Zhu, Zeyuan and Li, Yuanzhi and Wang, Shean and Wang, Lu and Chen, Weizhu},
  journal={arXiv preprint},
  year={2021}
}

@software{vonwerra2020trl,
  title   = {{TRL: Transformers Reinforcement Learning}},
  author  = {von Werra, Leandro and Belkada, Younes and Tunstall, Lewis and Beeching, Edward and Thrush, Tristan and Lambert, Nathan and Huang, Shengyi and Rasul, Kashif and Gallouédec, Quentin},
  license = {Apache-2.0},
  year    = {2020}
}

@misc{hu2024agentscodriverlargelanguagemodel,
      title={{A}gents{C}o{D}river: {L}arge {L}anguage {M}odel {E}mpowered {C}ollaborative {D}riving with {L}ifelong {L}earning},
      author={Senkang Hu and Zhengru Fang and Zihan Fang and Yiqin Deng and Xianhao Chen and Yuguang Fang},
      year={2024},
      eprint={2404.06345},
      archivePrefix={arXiv},
      primaryClass={cs.AI},
  journal = {arXiv preprint},
}

@ARTICLE{10976336,
  author={Hu, Senkang and Fang, Zhengru and Fang, Zihan and Deng, Yiqin and Chen, Xianhao and Fang, Yuguang and Kwong, Sam Tak Wu},
  journal={IEEE Transactions on Mobile Computing},
  title={{A}gents{C}o{M}erge: {L}arge {L}anguage {M}odel {E}mpowered {C}ollaborative {D}ecision {M}aking for {R}amp {M}erging},
  year={2025},
  volume={24},
  number={10},
  pages={9791-9805},
  doi={10.1109/TMC.2025.3564163}
}

@article{hu2026distributionaligned,
  title={{D}istribution-{A}ligned {D}ecoding for {E}fficient {L}lm {T}ask {A}daptation},
  author={Hu, Senkang and Han, Xudong and Jiang, Jinqi and Tao, Yihang and Fang, Zihan and Dai, Yong and Kwong, Sam and Fang, Yuguang},
  journal={Advances in Neural Information Processing Systems},
  volume={38},
  pages={56153--56188},
  year={2026}
}

@article{chen2025test,
  title={{T}est-{T}ime {D}istillation for {C}ontinual {M}odel {A}daptation},
  author={Chen, Xiao and Huang, Jiazhen and Liu, Zhiming and Jiang, Qinting and Huang, Fanding and Jiang, Jingyan and Wang, Zhi},
  journal={arXiv preprint},
  year={2025}
}

@article{huang2026gui,
  title={{G}{U}{I} {A}gents for {C}ontinual {G}ame {G}eneration},
  author={Huang, Yixu and Li, Bo and Li, Na and Wang, Zhe and Chen, Kaijie and Ge, Haonan and Si, Qingyi and Shen, Yuanzhe and Yang, Ruihan and Wang, Guangjing and others},
  journal={arXiv preprint},
  year={2026}
}

@article{ma2026thinking,
  title={{T}hinking with {B}lueprints: {A}ssisting {V}ision-{L}anguage {M}odels in {S}patial {R}easoning via {S}tructured {O}bject {R}epresentation},
  author={Ma, Weijian and Sun, Shizhao and Yu, Tianyu and Wang, Ruiyu and Chua, Tat-Seng and Bian, Jiang},
  journal={arXiv preprint},
  year={2026}
}

@article{rlvr-emp,
  title={{D}oes {R}einforcement {L}earning {R}eally {I}ncentivize {R}easoning {C}apacity in {L}lms {B}eyond the {B}ase {M}odel?},
  author={Chen, Zhiqi and Lu, Rui and Zhao, Andrew and Wang, Zhaokai and Yue, Yang and Song, Shiji and Huang, Gao},
  journal={Advances in Neural Information Processing Systems},
  volume={38},
  pages={57654--57689},
  year={2026}
}

@inproceedings{ma2025instruct,
  title={{I}nstruct {W}here the {M}odel {F}ails: {G}enerative {D}ata {A}ugmentation via {G}uided {S}elf-{C}ontrastive {F}ine-{T}uning},
  author={Ma, Weijian and Chen, Ruoxin and Zhang, Keyue and Wu, Shuang and Ding, Shouhong},
  booktitle={Proceedings of the AAAI Conference on Artificial Intelligence},
  volume={39},
  number={6},
  pages={5991--5999},
  year={2025}
}

@article{liu2026all,
  title={{D}o {A}ll {I}ndividual {L}ayers {H}elp? {A}n {E}mpirical {S}tudy of {T}ask-{I}nterfering {L}ayers in {V}ision-{L}anguage {M}odels},
  author={Liu, Zhiming and Wei, Yujie and Feng, Lei and Su, Xiu and Xia, Xiaobo and Guan, Weili and Xie, Zeke and Yang, Shuo},
  journal={arXiv preprint},
  year={2026}
}

@article{li2026correcting,
  title={{C}orrecting {V}isual {B}lur {I}nduced by {A}ttention {D}istraction to {R}educe {H}allucinations: {A}lgorithm and {T}heory},
  author={Li, Quanjiang and Liu, Zhiming and Luo, Wei and Luo, Tingjin and Hou, Chenping},
  journal={arXiv preprint},
  year={2026}
}

@article{zhai2026rewards,
  title={{R}ewards as {L}abels: {R}evisiting {R}{L}{V}{R} from a {C}lassification {P}erspective},
  author={Zhai, Zepeng and Chen, Meilin and Zhao, Jiaxuan and Qian, Junlang and Shen, Lei and Lu, Yuan},
  journal={arXiv preprint},
  year={2026}
}

@article{su2026sell,
  title={{S}ell {M}ore, {P}lay {L}ess: {B}enchmarking {L}{L}{M} {R}ealistic {S}elling {S}kill},
  author={Su, Xuanbo and Hu, Wenhao and Su, Haibo and Chen, Yunzhang and Zhan, Le and Yang, Yanqi and Huang, Leo},
  journal={arXiv preprint},
  year={2026}
}

@article{zhang2026memmark,
  title={{M}em{M}ark: {S}tate-{E}volution {A}ttribution {W}atermarking for {A}gent {L}ong-{T}erm {M}emory {S}ystems},
  author={Zhang, Haobo and Mao, Xutao and Dong, Guangyuan and Li, Ziwei and Su, Xuanbo and Chen, Kaijie and Yang, Jing and Lin, Zheng},
  journal={arXiv preprint},
  year={2026}
}

@article{liu2026automated,
  title={{A}utomated {O}ptimization {M}odeling via a {L}ocalizable {E}rror-{D}riven {P}erspective},
  author={Liu, Weiting and Wu, Han and Kuang, Yufei and Han, Xiongwei and Zhong, Tao and Feng, Jianfeng and Lu, Wenlian},
  journal={arXiv preprint},
  year={2026}
}

@article{liu2026toolanchor,
  title={{T}ool{A}nchor: {A}nchoring {C}ounterfactual {C}ontext to {B}oost {A}gentic {T}ool-{U}se {C}apability},
  author={Liu, Weiting and Bi, Jieyi and Zhou, Wanqi and Feng, Jianfeng and Ma, Yining and Han, Ai and Lu, Wenlian},
  journal={arXiv preprint},
  year={2026}
}

@article{shi2026androtmem,
  title={{A}ndro{T}{M}em: {F}rom {I}nteraction {T}rajectories to {A}nchored {M}emory in {L}ong-{H}orizon {G}{U}{I} {A}gents},
  author={Shi, Yibo and Li, Jungang and Zhang, Linghao and Dongfang, Zihao and Wu, Biao and Tao, Sicheng and Yan, Yibo and Qin, Chenxi and Liu, Weiting and Lin, Zhixin and others},
  journal={arXiv preprint},
  year={2026}
}

@article{zhang2026temporal,
  title={{T}emporal {G}ains, {S}patial {C}osts: {R}evisiting {V}ideo {F}ine-{T}uning in {M}ultimodal {L}arge {L}anguage {M}odels},
  author={Zhang, Linghao and Li, Jungang and Hei, Yonghua and Tao, Sicheng and Dai, Song and Yan, Yibo and Dongfang, Zihao and Liu, Weiting and Qin, Chenxi and Li, Hanqian and others},
  journal={arXiv preprint},
  year={2026}
}

@article{fu2026revisiting,
  title={{R}evisiting on-{P}olicy {D}istillation: {E}mpirical {F}ailure {M}odes and {S}imple {F}ixes},
  author={Fu, Yuqian and Huang, Haohuan and Jiang, Kaiwen and Liu, Jiacai and Jiang, Zhuo and Zhu, Yuanheng and Zhao, Dongbin},
  journal={arXiv preprint},
  year={2026}
}

@article{webcogreasoner,
  title={{W}eb-{C}og{R}easoner: {T}owards {M}ultimodal {K}nowledge-{I}nduced {C}ognitive {R}easoning for {W}eb {A}gents},
  author={Guo, Yuhan and Guo, Cong and Sun, Aiwen and He, Hongliang and Yang, Xinyu and Lu, Yue and Zhang, Yingji and Guo, Xuntao and Zhang, Dong and Liu, Jianzhuang and Duan, Jiang and Xiao, Yijia and Wen, Liangjian and Xu, Hai-Ming and Dai, Yong},
  journal={arXiv preprint},
  year={2026}
}

@article{a2eval,
  title={{A}2{E}val: {A}gentic and {A}utomated {E}valuation for {E}mbodied {B}rain},
  author={Zhang, Shuai and Hu, Jiayu and Chen, Zijie and Ding, Zeyuan and Zhang, Yi and Zhang, Yingji and Zhou, Ziyi and Liao, Junwei and Zhou, Shengjie and Dai, Yong and Lan, Zhenzhong and Ju, Xiaozhu},
  journal={arXiv preprint},
  year={2026}
}

@article{zhu2026unified,
  title={Unified Context Evolution for LLM Agents},
  author={Zhu, Zixuan and Hu, Yitong and Dai, Yong and Fang, Junfeng and Jiang, Chunyang and Hu, Senkang and Zhao, Yuzhi},
  journal={arXiv preprint arXiv:2606.02304},
  year={2026}
}

\newpage
\appendix
\section{Artifact License}
\label{app:artifact_license}

The research artifacts associated with this paper include the training and evaluation code, configuration files, prompt templates, and the generated skill-bank files used by \method{}. 
These artifacts will be distributed for research use with MIT license in the incoming released repository. 
Users of the released artifacts should also comply with the licenses and terms of the underlying third-party resources, including the base Qwen3 models, the Qwen3 embedding model, the DAPO-Math-17K training data, the mathematical reasoning benchmarks, and the software libraries used for training and evaluation. 
The released skill banks are derived from training-time model trajectories and are intended to support reproduction and further research on skill-conditioned self-distillation.

\section{Use of AI Assistants}
\label{app:ai_assistants}

AI assistants were used during this project to help with code implementation, debugging, and language polishing. 
All core research ideas, method design choices, theoretical arguments, experimental designs, result interpretations, and final scientific claims were developed and checked by the authors. 
The authors reviewed and edited AI-assisted code and text before inclusion, and remain responsible for the correctness, integrity, and presentation of the paper and accompanying artifacts.

\section{Skill Bank Format}
\label{app:skill_bank}

\method{} uses a structured JSON skill bank. 
As shown in Tab. \ref{tab:skill_schema}, the implementation stores two knowledge fields: \texttt{general\_skills} and \texttt{common\_mistakes}. 
Each model scale uses a model-specific cold-start bank extracted from DAPO-Math training problems.
Fig.~\ref{fig:skill_bank_json} gives a concrete serialized example of this schema.
\begin{table*}[t]
\centering
\caption{JSON fields used by the \method{} skill bank.}
\label{tab:skill_schema}
\vspace{-0.8em}
\small
\setlength{\tabcolsep}{6pt}
\begin{tabular}{p{0.22\textwidth}p{0.70\textwidth}}
\toprule
Field & Meaning \\
\midrule
\texttt{general\_skills} & Reusable positive guidance. Each entry contains \texttt{skill\_id}, \texttt{title}, \texttt{principle}, and \texttt{when\_to\_apply}. \\
\texttt{common\_mistakes} & Reusable negative guidance. Each entry contains \texttt{mistake\_id}, \texttt{description}, \texttt{why\_it\_happens}, and \texttt{how\_to\_avoid}. \\
\texttt{metadata} & Provenance and merge statistics, including source description, merge group size, stagnation patience, and layer-level merge counts. \\
\bottomrule
\end{tabular}
\end{table*}

\begin{figure*}[t]
\centering
\begin{innerjson}
{
  "general_skills": [
    {
      "skill_id": "gen_001",
      "title": "Translate Constraints to Algebra",
      "principle": "Convert stated constraints into algebraic relations.",
      "when_to_apply": "When variables are governed by explicit conditions."
    }
  ],
  "common_mistakes": [
    {
      "mistake_id": "err_001",
      "description": "Skipping the constraint model.",
      "why_it_happens": "The solver starts computing before formalizing conditions.",
      "how_to_avoid": "Restate the configuration before deriving equations."
    }
  ],
  "metadata": {
    "source": "hierarchical merge from raw candidates",
    "merge_group_size": 32,
    "merge_stagnation_patience": 3
  }
}
\end{innerjson}
\caption{Example JSON instance for the \method{} skill bank. The line breaks follow the actual serialized structure but use shortened entry text for readability.}
\label{fig:skill_bank_json}
\end{figure*}

In the released artifacts used by our experiments, the Qwen3-1.7B, Qwen3-4B, and Qwen3-8B banks contain 28/24/46 general skills and 98/15/8 common mistakes, respectively. 

\section{Prompt and Skill Construction Details}
\label{app:prompts}

This section gives the prompt templates and algorithms used to construct and consume the skill bank. 
The templates are rendered as plain user messages before the model-specific chat template is applied.

\subsection{Student and Teacher Contexts}
\label{app:student_teacher_prompts}

\method{} keeps a strict information boundary. 
The student rollout and evaluation prompt contain only the problem, while each teacher prompt receives one retrieved skill--mistake guidance block in addition to the problem. 
Vanilla \method{} setting does not include a reference solution in the teacher prompt.
Fig.~\ref{fig:student_teacher_prompts} shows the corresponding student and teacher prompt templates.

\begin{figure*}[t]
\centering
\begin{minipage}{\textwidth}

\begin{promptcard}{Student Prompt}

\small

\begin{verbatim}
Problem: {problem}

Please reason step by step, and put your final answer within \boxed{}.
\end{verbatim}

\end{promptcard}

\vspace{0.8em}

\begin{promptcard}{Teacher Prompt for Pair k}

\small

\begin{verbatim}
You may use the following retrieved math-reasoning guidance as soft guidance.
Solve the current problem independently and do not quote it verbatim.

### General Principles
- **{skill.title}**: {skill.principle}
  _Apply when: {skill.when_to_apply}_

### Mistakes to Avoid
- **Don't**: {mistake.description}
  **Instead**: {mistake.how_to_avoid}

Problem: {problem}

Please reason step by step, and put your final answer within \boxed{}.
\end{verbatim}

\end{promptcard}

\end{minipage}

\caption{
Student and teacher prompt templates in \method{}.
The student receives only the problem,
while each teacher scores the same student rollout
under one retrieved skill--mistake context.
}
\label{fig:student_teacher_prompts}

\end{figure*}

\subsection{Cold-Start Memory Generation and Skill Extraction}
\label{app:cold_start}

The cold-start pipeline first samples in-domain problems, generates model completions, scores them with the answer verifier, and converts them into compact memories. 
Successful memories are used to extract general skills, while failed memories are used to extract common mistakes.
Alg.~\ref{alg:cold_start} summarizes the construction procedure.
The three prompt templates used in this stage are shown in Tabs.~\ref{tab:prompt_memory_generation}, \ref{tab:prompt_success_skill}, and~\ref{tab:prompt_failure_skill}, respectively.

\begin{algorithm}[t]
\small
\caption{Cold-start skill construction}
\label{alg:cold_start}
\begin{algorithmic}[1]
\Require Seed set $\mathcal{D}_0$, base model $\pi_{\theta_0}$, verifier, extraction prompts
\State Initialize memory set $\mathcal{M}\leftarrow\emptyset$
\For{each problem $x\in\mathcal{D}_0$}
\State Render the memory-generation prompt and sample a completion from $\pi_{\theta_0}$
\State Extract the boxed answer and compute the verifier reward
\State Build a memory record with problem, completion, reward, answer, trajectory summary, and feedback
\State Add the memory record to $\mathcal{M}$
\EndFor
\State Split $\mathcal{M}$ into successful memories $\mathcal{M}^{+}$ and failed memories $\mathcal{M}^{-}$
\For{each memory $m\in\mathcal{M}^{+}$}
\State Extract up to three general-skill candidates with fields \texttt{title}, \texttt{principle}, and \texttt{when\_to\_apply}
\EndFor
\For{each memory $m\in\mathcal{M}^{-}$}
\State Extract up to three common-mistake candidates with fields \texttt{description}, \texttt{why\_it\_happens}, and \texttt{how\_to\_avoid}
\EndFor
\State Merge and re-index the two candidate collections using Alg.~\ref{alg:skill_merge}
\State Return the initialized skill bank $\mathcal{B}=\{\mathcal{G},\mathcal{E}\}$
\end{algorithmic}
\end{algorithm}
\begin{table*}[t]
\centering
\begin{promptcard}{Prompt: Cold-Start Memory Generation}
\small \ttfamily
You are a careful mathematical problem-solving agent.\par
\vspace{0.25em}
Solve the following problem step by step. Keep the reasoning coherent and self-contained, and end with a single final answer enclosed in \textbackslash boxed\{\{\}\}.\par
\vspace{0.25em}
Problem:\par
\{problem\}
\end{promptcard}
\caption{Prompt template for generating cold-start mathematical reasoning memories.}
\label{tab:prompt_memory_generation}
\end{table*}

\begin{table*}[t]
\centering
\begin{promptcard}{Prompt: General-Skill Extraction from a Successful Memory}
\small \ttfamily
You are an expert at distilling mathematical reasoning behavior into concise, reusable skills for a reinforcement-learning agent.\par
\vspace{0.25em}
You will be given ONE successful math problem-solving memory. The memory contains the original problem, a compact reasoning trajectory, and a summarized raw attempt.\par
\vspace{0.25em}
Your task:\par
1. Derive 1-3 GENERAL skills that likely contributed to the success.\par
2. Each skill must be broadly reusable across algebra, geometry, number theory, combinatorics, and olympiad-style reasoning.\par
3. Phrase each skill as an actionable principle; avoid task-specific constants, entity names, or one-off details unless they express a general method.\par
4. Merge overlapping ideas inside this response; do not output near-duplicate skills.\par
5. Use only evidence grounded in the provided memory.\par
\vspace{0.25em}
Successful memory:\par
\{memory\_json\}\par
\vspace{0.25em}
Return ONLY valid JSON with key \texttt{general\_skills}.
\end{promptcard}
\caption{Prompt template for extracting reusable positive skills from successful cold-start memories.}
\label{tab:prompt_success_skill}
\end{table*}

\begin{table*}[t]
\centering
\begin{promptcard}{Prompt: Common-Mistake Extraction from a Failed Memory}
\small \ttfamily
You are an expert at analyzing failed mathematical reasoning and turning failures into concise, reusable cautionary skills for a reinforcement-learning agent.\par
\vspace{0.25em}
You will be given ONE failed math problem-solving memory. The memory contains the original problem, summarized failure evidence, and the raw final attempt.\par
\vspace{0.25em}
Your task:\par
1. Derive 1-3 COMMON mistakes that explain the failure.\par
2. Each item must describe a general failure mode, why it happens, and how to avoid it in future math reasoning.\par
3. Make every item broadly reusable across algebra, geometry, number theory, combinatorics, and olympiad-style reasoning.\par
4. Merge overlapping ideas inside this response; do not output near-duplicate mistakes.\par
5. Use only evidence grounded in the provided memory.\par
\vspace{0.25em}
Failed memory:\par
\{memory\_json\}\par
\vspace{0.25em}
Return ONLY valid JSON with key \texttt{common\_mistakes}.
\end{promptcard}
\caption{Prompt template for extracting reusable mistake patterns from failed cold-start memories.}
\label{tab:prompt_failure_skill}
\end{table*}

\subsection{Skill Bank Merging}
\label{app:skill_merge}

Raw candidate skills are noisy and redundant, so the implementation merges each collection hierarchically. 
Each merge call receives up to 32 items, preserves unique insights, and returns a compact JSON collection. 
Merging stops when the root group is reached or when the item count stops decreasing for three consecutive layers. 
Alg.~\ref{alg:skill_merge} gives the merge procedure, while Tabs.~\ref{tab:prompt_merge_general} and~\ref{tab:prompt_merge_mistake} give the corresponding merge prompts.

\begin{algorithm}[t]
\small
\caption{Hierarchical skill merging}
\label{alg:skill_merge}
\begin{algorithmic}[1]
\Require Candidate collection $\mathcal{C}$, prompt type $q$, group size $H=32$, stagnation patience $P=3$
\State Set layer index $\ell\leftarrow0$ and stagnant counter $s\leftarrow0$
\Repeat
\State Partition $\mathcal{C}$ into groups of at most $H$ items
\For{each group with more than one item}
\State Render the merge prompt for type $q$ and parse the returned JSON
\State Fall back to the original group if parsing fails
\EndFor
\State Concatenate all merged groups into $\mathcal{C}'$
\State Update $s\leftarrow0$ if $|\mathcal{C}'|<|\mathcal{C}|$, otherwise $s\leftarrow s+1$
\State Set $\mathcal{C}\leftarrow\mathcal{C}'$ and $\ell\leftarrow\ell+1$
\Until{there is one group or $s\ge P$}
\State Remove exact duplicates and reassign IDs with prefix \texttt{gen} or \texttt{err}
\State Return the merged collection
\end{algorithmic}
\end{algorithm}
\begin{table*}[t]
\centering
\begin{promptcard}{Prompt: General-Skill Merging}
\small \ttfamily
You are an expert at consolidating independently-generated math skills into a compact, non-redundant skill bank.\par
\vspace{0.25em}
You will be given up to 32 general skills extracted from different memories. Some are duplicates, some partially overlap, and some are unique.\par
\vspace{0.25em}
Your task:\par
1. Merge semantically duplicate or strongly overlapping skills.\par
2. Preserve all unique insights.\par
3. Prefer the most general, transferable wording.\par
4. Treat recurrence as evidence that the pattern is systematic and synthesize one stronger skill.\par
5. Do not force a fixed final count.\par
6. Do not mention specific problems, source memories, or dataset names.\par
\vspace{0.25em}
General skills to merge:\par
\{items\_json\}\par
\vspace{0.25em}
Return ONLY valid JSON with key \texttt{general\_skills}.
\end{promptcard}
\caption{Prompt template for hierarchically merging general-skill candidates.}
\label{tab:prompt_merge_general}
\end{table*}

\begin{table*}[t]
\centering
\begin{promptcard}{Prompt: Common-Mistake Merging}
\small \ttfamily
You are an expert at consolidating independently-generated math failure lessons into a compact, non-redundant caution bank.\par
\vspace{0.25em}
You will be given up to 32 common-mistake items extracted from different memories. Some are duplicates, some partially overlap, and some are unique.\par
\vspace{0.25em}
Your task:\par
1. Merge semantically duplicate or strongly overlapping mistakes.\par
2. Preserve all unique insights.\par
3. Prefer the most general, transferable wording.\par
4. Treat recurrence as evidence that the failure pattern is systematic and synthesize one stronger mistake item.\par
5. Do not force a fixed final count.\par
6. Do not mention specific problems, source memories, or dataset names.\par
\vspace{0.25em}
Common mistakes to merge:\par
\{items\_json\}\par
\vspace{0.25em}
Return ONLY valid JSON with key \texttt{common\_mistakes}.
\end{promptcard}
\caption{Prompt template for hierarchically merging common-mistake candidates.}
\label{tab:prompt_merge_mistake}
\end{table*}

\subsection{Online Skill Evolution}
\label{app:skill_evolution}

The online updater reuses the same success/failure extraction prompts and merge prompts. 
It updates only the dynamic portion of the bank; static cold-start entries are preserved.
Alg.~\ref{alg:skill_evolution} gives the complete online update procedure.

\begin{algorithm}[t]
\small
\caption{Online skill evolution (update and maintenance)}
\label{alg:skill_evolution}
\begin{algorithmic}[1]
\Require Recent rollout records $\mathcal{W}$, current bank $\mathcal{B}$, update frequency $F$, success threshold $\gamma$
\Require Max new items $N$, dynamic capacity $C$
\If{online update is disabled, $\mathcal{W}$ is empty, or the current step is not divisible by $F$}
\State Return $\mathcal{B}$
\EndIf
\State Gather rollout records across workers and compute success rate $\hat{p}$
\If{$\hat{p}\ge\gamma$}
\State Return $\mathcal{B}$
\EndIf
\State Split $\mathcal{W}$ into successful records $\mathcal{W}^{+}$ and failed records $\mathcal{W}^{-}$
\State Extract new general-skill candidates from $\mathcal{W}^{+}$
\State Extract new common-mistake candidates from $\mathcal{W}^{-}$
\State Merge new candidates of each type
\State Merge the new dynamic candidates with existing dynamic entries
\State Keep at most $N$ net new dynamic entries per update and at most $C$ dynamic entries per collection
\State Replace the dynamic subset in $\mathcal{B}$ and save the latest bank and a step-indexed snapshot
\end{algorithmic}
\end{algorithm}

In our implementation, $F=25$, $\gamma=0.8$, $N=5$, and $C=30$ when dynamic skill updates are enabled. 
The updater can use either the current training model or a separate generation backend.

\section{Proof Details}
\label{app:proof}

\subsection{Derivative of the Gated Loss}
\label{app:gate_derivative}

The gated token loss is
\begin{equation}
    \ell_{\mathrm{gate}}(\Delta)
=
\log 2-
\log\left(1+\exp\left(-\frac{\Delta^2}{2\tau_g}\right)\right).
\end{equation}

Let $z(\Delta)=-\Delta^2/(2\tau_g)$, then
\begin{align}
\frac{\partial \ell_{\mathrm{gate}}(\Delta)}{\partial \Delta}
&=
-
\frac{\exp(z(\Delta))}{1+\exp(z(\Delta))}
\frac{\partial z(\Delta)}{\partial \Delta}
\\
&=
\frac{\Delta}{\tau_g}
\frac{\exp(-\Delta^2/(2\tau_g))}
{1+\exp(-\Delta^2/(2\tau_g))}
\\
&=
\frac{\Delta}
{\tau_g\left(1+\exp(\Delta^2/(2\tau_g))\right)}.
\end{align}
This gives the gradient factor
\begin{equation}
g_\tau(\Delta)=
\frac{\Delta}
{\tau_g\left(1+\exp(\Delta^2/(2\tau_g))\right)}.
\end{equation}

\subsection{Shape of the Gated Objective}
\label{app:gate_shape}
\begin{figure*}[t]
\centering
\includegraphics[width=0.78\textwidth]{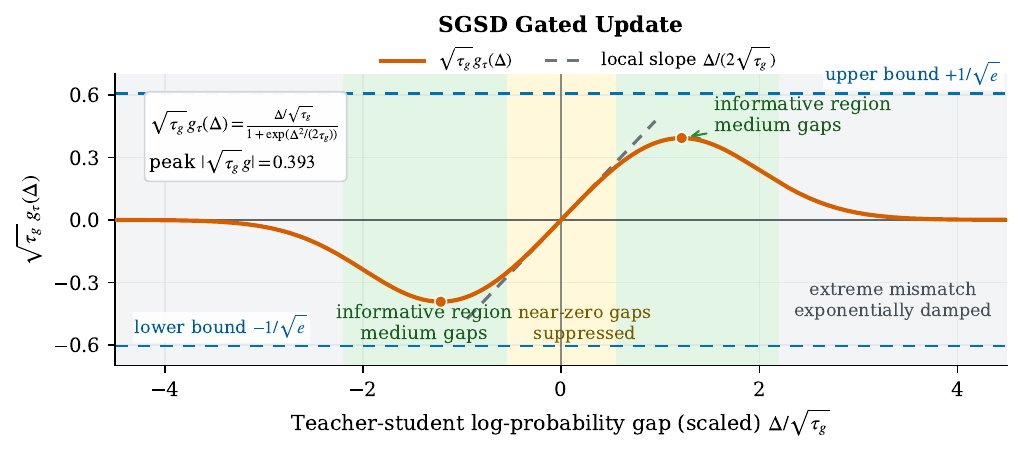}
\caption{\textbf{Shape of the gated objective.}}
\label{fig:gate_function}
\end{figure*}

As shown in Fig.~\ref{fig:gate_function}, the gated loss $\ell_{\mathrm{gate}}(\Delta)$ is symmetric and bounded: negligible teacher-student gaps receive little weight, while very large gaps saturate instead of growing without limit.
Its derivative $g_\tau(\Delta)$ preserves the sign of the log-probability gap $\Delta$ in the informative region, but decays toward zero for extreme gaps.
Thus, the gate keeps medium teacher-student disagreements as useful dense supervision while preventing outlier logits or prompt artifacts from dominating the update.

\subsection{Policy-Gradient Form}
\label{app:policy_gradient}

For teacher $k$ and token $t$, recall that
$
\Delta_t^{(k)}
=
\log p_T^{(k)}(y_t\mid x,y_{<t})
-
\log p_S(y_t\mid x,y_{<t}).
$
Since teacher logits are stop-gradient targets,
\begin{equation}
\nabla_\theta \Delta_t^{(k)}
=
-\nabla_\theta\log p_S(y_t\mid x,y_{<t}).
\end{equation}
Let $Z=\sum_t m_t+\epsilon$. 
For
$
\bar{\ell}^{(k)}
=
\frac{1}{Z}\sum_{t=1}^{T}m_t\ell_{\mathrm{gate}}(\Delta_t^{(k)}),
$
we have
\begin{equation}
\begin{aligned}
\small
\nabla_\theta \bar{\ell}^{(k)}
=&
-\frac{1}{Z}
\sum_{t=1}^{T}
m_t g_\tau(\Delta_t^{(k)})
\\
& \quad\cdot
\nabla_\theta\log p_S(y_t\mid x,y_{<t}).
\end{aligned}
\end{equation}
Substituting this into Eq.~\eqref{eq:loss-overall} yields
\begin{equation}
\begin{aligned}
\small
-\nabla_\theta \mathcal{L}_{\mathrm{\method{}}}(x)
&=
\sum_{k=1}^{K_x}\sum_{t=1}^{T}
\alpha_k(x)\rho_k
\frac{m_t}{Z}
g_\tau(\Delta_t^{(k)})
\\
&\quad\cdot
\nabla_\theta\log p_S(y_t\mid x,y_{<t}).
\end{aligned}
\end{equation}
Thus, \method{} is a policy-gradient-style update whose dense credit is controlled by teacher polarity and the gated gap derivative.

\subsection{Sign and Bounded Influence}
\label{app:bounded_gate}

For $\Delta\neq0$, the denominator of $g_\tau(\Delta)$ is positive, so
\begin{equation}
\operatorname{sign}(g_\tau(\Delta))=\operatorname{sign}(\Delta).
\end{equation}
The outcome-derived polarity $\rho_k$ therefore decides whether a teacher's token-level support is distilled or reversed. 
For boundedness,
\begin{equation}
1+\exp\left(\frac{\Delta^2}{2\tau_g}\right)
\ge
\exp\left(\frac{\Delta^2}{2\tau_g}\right),
\end{equation}
and hence
\begin{equation}
|g_\tau(\Delta)|
\le
\frac{|\Delta|}{\tau_g}
\exp\left(-\frac{\Delta^2}{2\tau_g}\right).
\end{equation}
The right-hand side is maximized at $|\Delta|=\sqrt{\tau_g}$, which gives
\begin{equation}
|g_\tau(\Delta)|\le \frac{1}{\sqrt{e\,\tau_g}}.
\end{equation}
Therefore, the token-level coefficient induced by the gate cannot grow without bound as teacher-student gaps become extreme.

\subsection{Local Approximation}
\label{app:approx}

Near zero, the expansion of $g_\tau$ is
\begin{equation}
g_\tau(\Delta)
=
\frac{\Delta}{2\tau_g}
 + O\left(\frac{\Delta^3}{\tau_g^2}\right),
\end{equation}
and the loss itself satisfies
\begin{equation}
\ell_{\mathrm{gate}}(\Delta)
=
\frac{\Delta^2}{4\tau_g}
 + O\left(\frac{\Delta^4}{\tau_g^2}\right).
\end{equation}
Thus negligible teacher-student gaps induce only small updates. 
For large gaps,
\begin{equation}
g_\tau(\Delta)
=
\frac{\Delta}{\tau_g}
\exp\left(-\frac{\Delta^2}{2\tau_g}\right)(1+o(1)),
\quad |\Delta|\to\infty,
\end{equation}
so extreme mismatches are exponentially damped.

Finally, consider locally close teacher and student distributions at a fixed history $h_t$, and define
\begin{equation}
\Delta_v=
\log p_T(v\mid h_t)-\log p_S(v\mid h_t).
\end{equation}
After the first-order normalization term cancels, reverse KL has the second-order expansion
\begin{equation}
D_{\mathrm{KL}}(p_T\,\|\,p_S)
=
\frac{1}{2}\sum_v p_T(v\mid h_t)\Delta_v^2
 + O(\|\Delta\|^3).
\end{equation}
Its local gradient is therefore linear in the log-probability gap. 
The gated objective matches this local linear behavior up to the constant scale $1/(2\tau_g)$, while deliberately departing from reverse KL for extreme gaps.

\section{Experimental Details}
\label{app:exp_details}

\subsection{Shared and Baseline Configurations}
\label{app:baseline_config}

The main experiments use Qwen3-1.7B, Qwen3-4B, and Qwen3-8B with LoRA adaptation. 
Training uses the English subset of DAPO-Math-17K, and evaluation is performed on AIME24, AIME25, and HMMT25. 
All trainable methods use learning rate $5\times10^{-6}$, LoRA rank $64$, LoRA alpha $128$, bfloat16 training, FlashAttention 2, and gradient checkpointing. 
Evaluation uses $12$ samples per problem, temperature $1.0$, top-$p$ $0.95$, no top-$k$ truncation, max new tokens $38912$, and symbolic answer checking with a string-based fallback.
Tab.~\ref{tab:exp_method_hparams} lists the method-specific hyperparameters that differ across baselines.

\begin{table*}[t]
\centering
\caption{Key training hyperparameters by method. Empty entries indicate that the parameter is not used by that method.}
\label{tab:exp_method_hparams}
\vspace{-0.8em}
\small
\setlength{\tabcolsep}{5pt}
\resizebox{\textwidth}{!}{%
\begin{tabular}{lccccc}
\toprule
Hyperparameter & OPSD & OPSD+Skill & GRPO & GRPO+Skill & \method{} \\
\midrule
Completion length & $1024$ & $1024$ & $16000$ & $16000$ & $1024$ \\
Prompt length & $20000$ & $20000$ & $2048$ & $2048$ & $20000$ \\
Temperature & $1.1$ & $1.1$ & $1.2$ & $1.2$ & $1.1$ \\
Top-$p$ / top-$k$ & $0.95$ / $20$ & $0.95$ / $20$ & -- & -- & $0.95$ / $20$ \\
Generations per prompt & $1$ & $1$ & $8$ & $8$ & $1$ \\
Teacher count per prompt & $1$ & $1$ & -- & -- & $8$ \\
Teacher update & fixed & fixed & -- & -- & live (synchronized) \\
Reference solution & yes & yes & -- & -- & no \\
\bottomrule
\end{tabular}%
}
\end{table*}

\subsection{\method{}-Specific Configuration}
\label{app:method_specific_config}

Tab.~\ref{tab:exp_method_specific_config} lists the \method{}-specific settings used by the final trainer. 
Named ablations change only the corresponding component, such as disabling token masking, removing polarity, using a single teacher, replacing the gated loss, or switching the teacher update strategy.

\begin{table*}[t]
\centering
\caption{\method{}-specific hyperparameters.}
\label{tab:exp_method_specific_config}
\vspace{-0.8em}
\small
\setlength{\tabcolsep}{6pt}
\begin{tabular}{p{0.28\textwidth}p{0.64\textwidth}}
\toprule
Component & Setting \\
\midrule
Cold-start bank & $256$ in-domain training problems per model scale \\
Retrieval encoder & Qwen3-Embedding-0.6B \\
Retrieved guidance & top-$8$ general skills and top-$8$ common mistakes \\
Teacher pool & one rank-aligned skill--mistake pair per teacher \\
Teacher weighting & softmax-normalized retrieval scores \\
Teacher policy & live synchronized self-teacher \\
Privileged information & skill--mistake guidance only; no reference solution \\
Vocabulary support & full-vocabulary normalization \\
Gated objective & $\tau_g=1.0$ \\
Robust support estimation & token masking, gap clipping with $c_\Delta=3.0$, confidence threshold $\epsilon_a=0.05$ \\
Online skill update & $F=25$, $\gamma=0.8$, $N=5$, $C=30$ \\
\bottomrule
\end{tabular}
\end{table*}

\end{document}